\documentclass{article}
\usepackage{times}  
\usepackage{helvet}  
\usepackage{courier}  
\usepackage[hyphens]{url}  
\usepackage{graphicx} 
\usepackage{natbib}   
\usepackage{caption}  

\usepackage{algorithmic}

\usepackage{placeins}
\usepackage{authblk}
\usepackage{fancyhdr}
\usepackage[utf8]{inputenc}
\usepackage[T1]{fontenc}   
\usepackage{booktabs}       
\usepackage{amsfonts}     
\usepackage{nicefrac}      
\usepackage{xspace}      
\usepackage{dsfont}
\usepackage{aliascnt}
\usepackage{amsfonts,amsmath,amssymb,amsthm}
\usepackage{multirow}
\usepackage{mathrsfs}
\usepackage{tabularx}
\usepackage{standalone}
\usepackage{stmaryrd}
\usepackage{textcomp}
\usepackage{subcaption}
\usepackage[ruled,vlined]{algorithm2e}
\usepackage{algorithmic}
\usepackage{diagbox}
\usepackage{longtable}

\newcommand{\Voc}{\mathcal{V}}
\newcommand{\TVoc}{\mathcal{V}^{-}}
\newcommand{\params}{\theta}
\newcommand{\policy}{\pi_{\params}}
\newcommand{\policyold}{\pi_{\params_{old}}}
\newcommand{\exploration}{\pi^{-}_{\params}}
\newcommand{\explorationold}{\pi^{-}_{\params_{old}}}
\newcommand{\image}{\mathcal{I}}
\newcommand{\answer}{\mathcal{A}}

\newcommand{\algo}{TrufLL\xspace}
\newcommand{\algolong}{TRUncated ReinForcement Learning for Language\xspace}{}
\newcommand{\indicatrice}[1]{\mathds{1}_{#1}}

\DeclareMathOperator{\clip}{clip}
\DeclareMathOperator*{\argmin}{argmin}   
\DeclareMathOperator*{\argmax}{argmax} 

\DeclareMathOperator{\topk}{top(k)}
\newcommand{\topkarg}[1]{\operatorname{top(k=#1)}}
\DeclareMathOperator{\topp}{top(p)}
\newcommand{\topparg}[1]{\operatorname{top(p=#1)}}
\DeclareMathOperator{\pth}{p_{th}(\alpha)}
\newcommand{\ptharg}[1]{\operatorname{p_{th}(\alpha=#1)}}
\DeclareMathOperator{\sample}{sample(k)}
\newcommand{\samplearg}[1]{\operatorname{sample(k=#1)}}

\title{Learning Natural Language Generation from Scratch with Truncated Reinforcement Learning}
\date{}

\author[$\dag$, $\ddag$]{Alice Martin}
\author[$\dag$]{Guillaume Quispe}
\author[$\dag$]{Charles Ollion}
\author[$\ddag$]{Sylvain Le Corff}
\author[$\top$]{Florian Strub}
\author[$\star$]{Olivier Pietquin}

\affil[$\dag$]{{\small CMAP, \'Ecole Polytechnique, Institut Polytechnique de Paris, Palaiseau.}}
\affil[$\ddag$]{{\small Samovar, T\'el\'ecom SudParis, d\'epartement CITI, TIPIC, Institut Polytechnique de Paris, Palaiseau.}}
\affil[$\top$]{{\small Deep Mind.}}
\affil[$\star$]{{\small Google Research, Brain Team.}}

\usepackage{geometry}
\begin{document}

\maketitle

\begin{abstract}
This paper introduces \algolong (\algo), an original approach to train conditional language  models from scratch by only  using reinforcement learning (RL). 
As RL methods unsuccessfully scale to large action spaces, we dynamically truncate the vocabulary space using a  generic language model. 
\algo thus enables to train a language agent by solely interacting with its environment without  any task-specific prior knowledge; it is only guided with a task-agnostic language model. Interestingly, this approach avoids the dependency to labelled datasets and inherently reduces pretrained policy flaws 
such as language or exposure biases. 
We evaluate \algo on two visual question generation tasks, for which we report positive results over performance and language metrics, which we then corroborate with a human evaluation. 
To our knowledge, it is the first approach that successfully learns a language generation policy (almost) from scratch. 
\end{abstract}

\section{Introduction}
\label{sec:introduction}
Since the development of generic language models trained on massive unlabelled text corpora~\citep{radford2019language, brown2020language}, state-of-the art language processing systems rely on \textit{sequential transfer learning}~\citep{ruder2019neural}. The pretrained Language Model (LM) is fine-tuned on the downstream task using a standard supervised learning (SL) objective~\citep{wu2019alternating,peters2019tune}.
Yet, such an approach suffers from several issues~\citep{chen2020recall}:
(i) catastrophic forgetting when a model forgets previously learned knowledge and overfits to target domains, (ii) computational inefficiency from fine-tuning billion-parameters networks, and (iii) the need of supervised datasets. 
Moreover, task-specific language models learned with SL suffer from well-studied text degeneration issues~\citep{holtzman2019curious}, such as the exposure bias \citep{bengio2015scheduled}, language biases~\citep{saleh2020hierarchical,jaques2020human},
or a lack of diversity~\citep{li2015diversity}. 
\begin{figure}[t]
\includegraphics[scale=0.8]{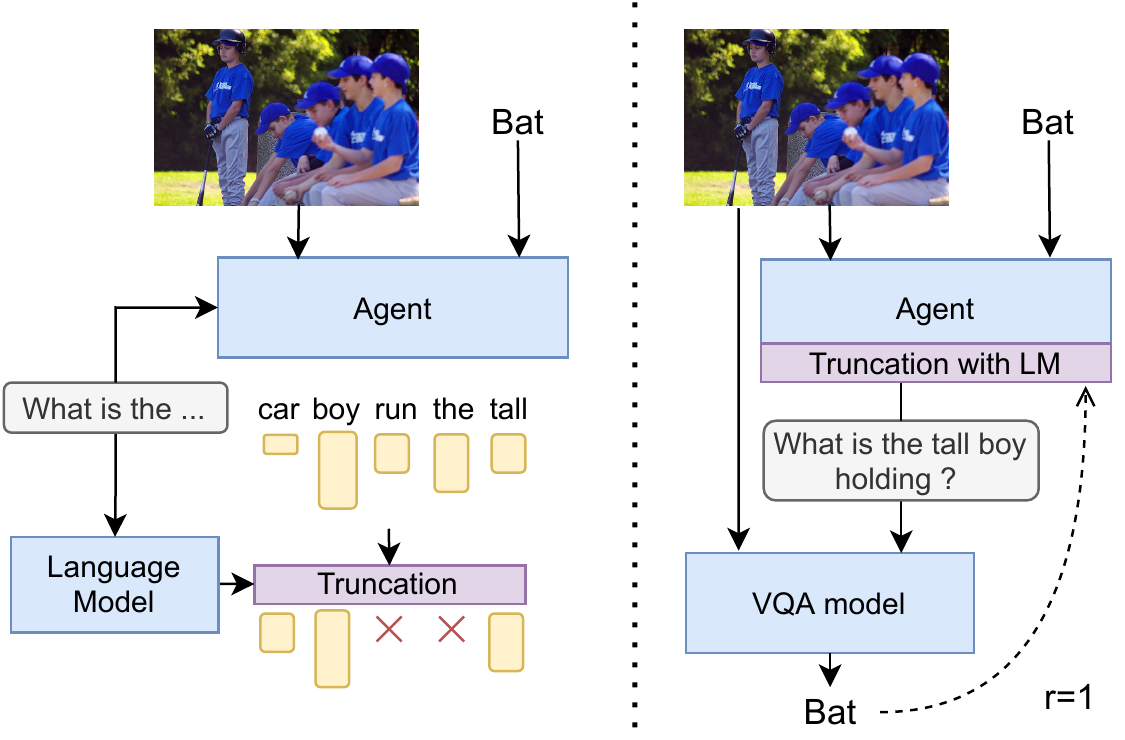}
\centering
\caption{\small (left) In a conditional language generation task as VQG, \algo truncates the vocabulary space by using a language model. Here, 'run,' and 'the' are syntactically incorrect and thus truncated. Yet, 'car' is not trimmed as the LM is not visually grounded. (right) In a VQG training loop, the agent generates a question given an image-answer pair, which is then fed to a VQA model predicting an expected answer. If both answers match, the agent is rewarded.}
\label{fig:truffl_sketch}
\vskip -1em 
\end{figure}

On the other hand, text generation can be naturally framed as a sequential decision making problem, with the sequence of words seen as successive actions over a vocabulary. Thus, some researchers have recently focused on learning language models using instead Reinforcement Learning (RL)~\citep{strub2017end,das2017visual,narasimhan2015language}. 
Like human learning, RL methods allow acquiring language through interactions within rich and diverse environments. 
In addition, they allow optimizing a non-differentiable learning signal and thus to handle a more diverse set of objective functions. Finally, they avoid some of the text degeneration issues previously mentioned.
So far RL-based text-generation tasks 
have relied on a pre-training phase to ease learning: the policy language model is trained with SL on the task dataset, before being fine-tuned with policy gradient methods~\citep{sutton1999policy} on the task at hand. Those approaches often require human-labelled datasets. Besides, combining pre-training and fine-tuning phases
either barely change the policy distribution,  
or induces \textit{language drift}~\citep{lazaridou2020multi,lu2020countering}, 
i.e the generated language drifts semantically or syntactically from natural language.

In this paper, we aim at learning a conditional language model using \textit{RL from scratch}, so that (i) we get free from datasets with human annotations, and (ii) we avoid the text generation flaws induced by the common methods. While appealing, such an approach requires overcoming the hurdle of the combinatorial language action space, a vocabulary usually containing more than 10,000 words.
Yet, while large and discrete, a language action space contains a specific structure, made of all the grammatical, syntactical, and semantics rules of a given language.
\algo leverages such structure to drive the exploration of the RL-based language agent during training. At each time step of the text generation process, \algo truncates its effective action space to a small subset of words provided by a pretrained task-agnostic language model. 
Such an approach injects a generic prior linguistic knowledge into the RL algorithm, is usable on tasks lacking in-domain labeled data, and can be easily transferred to new RL-based text generation tasks. Thus, \algo can be applied to any language generation task given a generic LM and a reward.
We here evaluate it on two Visual Question Generation (VQG) tasks, the synthetic CLEVR dataset (Johnson et al. 2017), and the natural language VQAv2 dataset (Goyal et al. 2017). Unlike alternative RL from scratch approaches, \algo manages to ask meaningful and valid questions, exhibiting success rate and language metrics close to classic pretrain models with labeled data. It also produces original language which differs from the dataset initial distribution, while scaling to large vocabularies containing over 10,000 words.

\section{Background}
\label{sec:background}
\paragraph{Language Generation as an RL Problem.}
We cast the word-based text generation task as a Markov Decision Process to apply RL methods~\cite{sutton1998introduction}. In this setting, a language model's agent generates a sequence of words $w_{<t} = (w_0, w_1, \dots, w_{t-1})$ drawn from a vocabulary $\Voc$, given an initial context $c$ associated with a reward $r_t$. 
Translation, text summarization or image captioning are examples of such tasks respectively using a source sentence, a text article, or an image as a context ($c$). 
During this process, the agent may be rewarded with language scores~\citep{ranzato2015sequence}, human preferences~\citep{stiennon2020learning} or task completion scores ~\citep{strub2017end}.

Formally, a language generation agent is defined by a policy $\policy$ (a distribution over $\Voc$) parametrized by $\params$, first initialized with the context $c$. At each time step $t$, the agent samples a new word $w_t$ from its policy $\policy(w_t|w_{<t}, c)$. It moves to a new state $(w_{<t+1}, c)$ and receives a reward $r_t=r(w_{<{t}}, c, w_{t})$, where $r(.)$ is a reward function relative to the language task. The RL language agent aims to learn a policy that maximizes  $\mathbb{E}_{\policy}[\sum^T_{t=0} r_t]$,\footnote{We cast the language modelling as an episodic problem with $\gamma=1$ and omit the discount factor in the paper for clarity.} while generating the sequence of words $w_{<T}$, where $\mathbb{E}_{\policy}$ is the expectation under $\policy$, and $T$ the maximal length of the words sequence.

\paragraph{Policy Gradient} This optimization process may be performed through Policy Gradient (PG) algorithms~\citep{sutton1999policy}. 
In the language literature, REINFORCE~\citep{williams1992simple} has been used as a simple Monte Carlo approximation of this gradient ~\citep{strub2017end,li2016deep}.
Yet, in this paper, we use a Proximal Policy Optimization approach (PPO)~\citep{schulman2017proximal} to have a lower variance and better convergence rate; PPO clips the gradient estimate to have smooth policy updates.

Given a state $s_t=(w_{<t}, c)$ and an action $a_t=w_t$, PG methods minimize the objective $L_{PG} =\mathbb{\widehat{E}}_{t}[\log \policy(a_t|s_t) \hat{A}_t]$ where $\hat{A}$ is an estimator of the advantage function, here defined as $\hat{A}_t= r_t - V(s_t)$ with $V(s)=\mathbb{E}_{\policy}[\sum_t r(s_t, a_t)|s]$. PPO then keeps track of the previous policy $\policyold$ before the PG update to compute the training objective:
\begin{equation*}
    L_{PPO} = \mathbb{E}_{\policy}\big[\min (\rho_t A_t, \clip(1-\epsilon, \rho_t, 1 + \epsilon)A_t)\big]\,,
\end{equation*}  
where $\rho_t=\policy(a_t|s_t)/\policyold(a_t| s_t)$ and $\epsilon$ is a hyper-parameter controlling the magnitude of the policy updates. Finally, the training loss is completed with (i) a value-based loss to learn a baseline that reduces the gradient variance, here $L_{VF} = \vert \sum_{t=i}^{T} r_i - V(s_t) \vert^{2}_{2}$,
\footnote{Note that other TD-based losses are applicable~\citep{sutton1998introduction,schulman2015high,espeholt2018impala}.}
(ii) an entropy term to soften the policy distribution $L_{E} = H(\policy(a_t|s_t))$.

\section{\algo}
\label{sec:truncation:model}

We here aim at making RL methods feasible in the language setting by dynamically reducing the action space, i.e., by restricting the language agent to select a word within a subset of the vocabulary at each time step. We detail below the action space's truncation model and the associated RL algorithm to learn the language agent. 

\subsection{Dynamic Vocabulary Truncation}
\label{subsec:trufll:truncation}
\algo combines two distinct language models, which share the same vocabulary $\Voc$: a RL language agent $\policy$ and a pretrained language model $f_{LM}$. 
At each timestep $t$, \algo restricts the vocabulary space of the RL language agent with: 
\vspace{-0.3em}
\begin{equation*}
  \TVoc_t = \{ w | w \in \Voc, g_{trunc}(w|w_{<t}) =1 \}\,,
\end{equation*}
where $g_{trunc}$ is a truncation function based on $f_{LM}$ which either associates 0 or 1 with each word in the vocabulary given the past words $w_{<t}$. From a language modelling perspective, the vocabulary space of the language agent is reduced from $\Voc$ to $\TVoc$ where $|\TVoc| \ll |\Voc|$, with $|\cdot|$ the cardinal of a finite set. From a RL perspective, the RL agent follows a truncated policy $\policy^{-}$ which only samples actions over the subset $\TVoc$. In practice,  
such a policy is computed using a masked softmax function over the truncated vocabulary $\TVoc_t$:
$\policy^{-}(.|w_{<t}, c)=\mathrm{softmax}(m*logits_{\policy}(w_{<t}, c))$ where $m=1$ when $g_{trunc}(w|w_{<t})=1$ otherwise $m=-\infty$.

\subsection{Truncation Functions}
We here list the different truncation functions $g_{trunc}$ explored through the paper.

\paragraph{Top-k words:}
This function selects the $k$ words with the highest probability given by $f_{LM}(.|w_{<t})$:
\vspace{-0.3em}
\begin{equation*}
    g_{\topk}(w_t|w_{<t}; k) = \indicatrice{w_t \in \topk(f_{LM}(.|w_{<t}))}\,.
\end{equation*}

\paragraph{Probability threshold ($\alpha$):} 
This function only keeps words having a probability $f_{LM}(.|w_{<t})$ greater than $\alpha$:
\vspace{-0.3em}
\begin{equation*}
g_{\pth}(w_t|w_{<t}; \alpha) = \indicatrice{f_{LM}(w_t|w_{<t}) > \alpha}\,.
\end{equation*}
    
\paragraph{Top-p:}
This function is based on nucleus sampling~\citep{holtzman2019curious}, and it keeps the most likely words contained in a probability mass $p$ of $f_{LM}(.|w_{<t})$. Formally, we define $\Voc_t^p$ as:
\vspace{-0.3em}
$$
     \Voc_t^p = \argmin_{|V_t|, V_t \subset \Voc}\{ w | w \in V_t, \sum_{w \in V_t} f_{LM}(w|w_{<t}) > p \}\,, 
$$
and readily, $g_{\topp}(w_t|w_{<t}; p) = \indicatrice{w_t \in \Voc_t^p}$.

\paragraph{Sample ($k$):}
This function randomly samples $k$ words from the language model with replacement to directly build the truncated vocabulary:
$$
g_{\sample}(w_t|w_{<t}; k) = \indicatrice{w_t \in \{w_i \sim f_{LM}(.|w_{<t})\; i \in \llbracket 1, \dots, k \rrbracket \}}\,.
$$
Only $\topk$ provides a fixed number of words at each time step. $\pth$, $\topp$, and $\sample$ have a dynamic truncation, whose size at $t$ depends on the language model entropy. 

\subsection{Task-Specific vs. Generic LM}
\label{subsec:trunc:LM}
We benchmark two types of language models for truncation. 
On the one hand, we use an \textit{external language model} pretrained on a large task-agnostic language corpora. Such a model provides a generic linguistic prior to the RL agent exploration process, solely encoding syntactic and semantic information. 
On the other hand, we use a \textit{task-related language model} pretrained on the supervised dataset associated with the task. Such a model provides a task-specific linguistic prior to the RL language agent, 
and captures language pragmatics. 
We emphasize that this paper aims at leveraging task-agnostic language models as they discard the need for task-specific data. For the sake of completeness, we also study the truncation with the task-related LM as an additional benchmark to assess our approach.

\section{Experimental Setting}
\label{sec:experiments}

We here list the experimental setting and detail the network and hyperparameters in Appendix~\ref{sec:app:hyperparamenters}.

\subsection{Visual Question Generation} 
\label{subsec:exp:VQG}

We showcase \algo on the task of Visual Question Generation (VQG)~\citep{mostafazadeh2016generating}, which is a form of Visual Jeopardy!~\texttrademark~\citep{ferrucci2012introduction}.
There, the language agent observes an image-answer pair and has to generate a question that results in a similar answer, as illustrated in  Figure~\ref{fig:truffl_sketch}. Such a task presents multiple advantages. First, by combining vision, scene understanding and language generation, it requires high-level reasoning and exhibits a large spectrum of language difficulties. Secondly, the success criterion is naturally non-differentiable, hence a natural fit for RL methods. Such a criterion, unlike metrics based on ground-truth sentences, allows generating diverse grounded questions given an image-answer pair.

Formally, the initial context $c$ is composed of the image-answer pair $(\image, \answer)$. The RL agent then generates a sequence of words $w_{<t}$ of maximum length $T$. We then provide the generated question to a pretrained VQA model. This model takes as inputs the image $\image$, the generated question $w_{<t}$ and outputs a predicted answer $\hat{\answer}$. Finally, the agent receives a reward $r(w_t, w_{<t}, c)$ based on $\answer$ and $\hat{\answer}$.

\subsection{Datasets} 
We evaluate \algo on the CLEVR and VQAv2 datasets to simulate large-scale VQG datasets. The two datasets have been originally created for the task of Visual Question Answering (VQA), i.e. for multi-modal classification algorithms predicting an answer given an image-question pair.

\paragraph{CLEVR}

The CLEVR VQA dataset~\citep{johnson2017clevr} is made of template questions on synthetic images, which contain simple objects with four distinct properties (shape, material, color, size). The vocabulary contains 86 words and 28 potential answers, making it a valuable proof of concept for assessing \algo. Both language models are single-layer LSTMs~\cite{hochreiter1997long} with 512 units, and 512 word embedding dimension. The task-specific LM is trained over the full train dataset of CLEVR questions. The external language model is trained on the mixture of CLOSURE~\cite{bahdanau2019closure} and CLEVR-Dialog \cite{kottur2019clevr} datasets. Although those two datasets share the CLEVR vocabulary, their language distribution differs from vanilla CLEVR. 
Finally, we use a pretrained GT-Vector-NMN~\citep{bahdanau2019closure} to compute the reward $r(w_t, w_{<t}, c) = \indicatrice{\answer=\hat{\answer}, t=T-1}$.

\paragraph{VQAv2}

The VQAv2 dataset~\citep{goyal2017making} is made of natural language and open-formed questions on images from the MS-Coco Dataset~\citep{lin2014microsoft}. It has a vocabulary of 14,810 words and 3,149 answers.

The task-specific language model is a one-layer LSTM with 512 units and a 512 word embedding dimension, pretrained over the full training dataset of VQAv2 questions. The External Language Model is Open-AI's GPT-2~\citep{radford2019language}. The original language model outputs a probability distribution over $50,257$ tokens, but we use a masked softmax function to restrict the probability distribution to the $14,810$ tokens of the VQAv2 dataset. Unlike most NLP tasks relying on pretrained generic language models, we do not fine-tune it on the task dataset. Instead, we leverage the few-shot generalization capabilities of GPT-2,  by feeding the language model with the prompt "\text{Here are a few examples:}" followed by 100 random questions $q_{<100}$ from the dataset (outside of the train set). The truncation is then based on 
the probability distribution $f^{gpt2}_{LM}(.|q_{<100}, w_{<t})$.
Finally, we used a pretrained VilBERT to compute the reward~\citep{Lu_2020_CVPR}.
Given the large number of answers, we use as reward a decreasing function of the rank of the reference answer $\mathrm{rk}(\answer)$:
$r(w_t, w_{<t}, c) = \indicatrice{\mathrm{rk}(\answer)\leq 10, t=T-1}e^{-\mathrm{rk}(\answer)/2}$, as further explained in Appendix~\ref{sec:app:reward}.

In these two settings, we acknowledge that the task dataset is still used to train the VQA models. Please note that the VQA modules are only used to model the environment, i.e. to provide a positive/negative feedback to the agent. In other settings, \algo would still work if we replace the VQA model by any language interface: text-game (e.g. Zork), expert-systems, or humans. Here, we only use the VQG framework as a proof of concept that natural language can be learned \emph{through pure interaction} given any task reward.

\subsection{Baselines}
\label{subsec:baseline} 

In this paper, we aim to show that a RL language agent can be trained from scratch, i.e. without the usual pre-training phase 
by solely interacting with another language system, the VQA model, when supported by truncation methods. 
The truncation with the task-related LM is referred to as \algo(Task-LM), while the one with the External LM is referred as \algo(Ext-LM). We first emphasize the difficulty of training an RL language agent from scratch through two baselines. We trained a simple on-policy PPO algorithm without any action space pruning, and refer to it as \textit{scratch}. Then, we added a KL regularization term to the loss, $L_{KL}=KL(\policy||f_{LM})$ and $L'=L+\lambda_{KL}L_{KL}$, to incorporate language prior to the agent as in ~\citep{jaques2017sequence,jaques2019way}.

We refer to it as \textit{scratch + KL-task} when distilling the task-specific language model, and \textit{scratch + KL-ext} with the external language model.
Finally, we include two baselines with a pre-training phase. We trained a language agent on the task-dataset with a log-likelihood objective, and refer to it as \textit{pretrain}. Then, we fine-tune the pretrained language agent with PPO without truncation, and refer to it as \textit{pretrain + RL fine-tune}. These two baselines should be viewed as gold standards as they rely on task-related data.

\subsection{Metrics and Evaluation Methods}
\label{subsec:evaluation-metrics}

Evaluating text generation is an open-research problem in language literature. 
We decompose automatic language evaluation into three categories to assess different facets of language, and perform as well a human evaluation study. 

\paragraph{Performance metrics.}
We measure the task-completion score or recall @ 1 which states whether the target answer $\answer$ is the top answer of the VQA models, and the recall @ 5 (R@5), which assesses whether $\answer$ is in the 5 top answers. These scores measure the task-solving abilities of the agent, but they are also conditioned by the VQA model abilities.

\paragraph{Language Metrics.}
First, we used n-grams metrics, BLEU \cite{papineni2002bleu}, METEOR \cite{banerjee2005meteor} and CIDEr \cite{vedantam2015cider}, to measure the similarity between the generated question and the reference questions in the evaluation set. While those scores can capture syntactic and semantic properties of language, they also fall short when dealing with open-form language, 
e.g. an identical answer may arise from two non-overlapping but syntactically correct questions. Thus, we also compute two metrics assessing the quality of the language independently of reference questions, the perplexity of the question given an external LM (ppl-e), and its perplexity given the task-related LM (ppl-t).

\paragraph{Diversity Metrics.} 
We here estimate a self-BLEU (sBLEU) score~\citep{zhang2017adversarial} over 10 questions generated on the same image-answer pair.
Although such score detects potential mode collapse, i.e., when the language utters identical sequences of words, it also values babbling, i.e., outputting random words. We thus also measure the probability mass of the ten most frequent words~\citep{choshen2020weaknesses}, and refer to it as peakiness (peak).

\paragraph{Human Evaluation.} On the VQAv2 task, we also performed human evaluation by surveying 53 participants on the first 50 questions produced by some of the models at test time. 
The study is based on pairwise comparison of question samples produced by the concurrent algorithms according to four criteria. First, we evaluated the language quality of the question samples, by asking the participants to select the most syntactically and semantically correct question among the two samples of the questions pair. Secondly, we evaluated language grounding, i.e adequacy of the sample to the image-answer pair, by asking the participants to select the question most suitable given the two elements. Thirdly, we evaluated the language originality and diversity, by asking participants to select the question the most different from the dataset reference question. Finally, we evaluated the number of syntax errors by asking participants to tick the question if it is grammatically incorrect.  Examples of questions asked during the study are included in the Appendix~\ref{app:human:eval}.

\subsection{Sampling methods for text generation}
\label{subsec:decoding:methods}

When generating text from a trained language model, the quality and diversity of samples depend on the decoding algorithm \citep{zhang2020trading}. We consider three text generation methods. \textit{greedy} uses the $\argmax$ of the policy, while \textit{sampling} uses the multinomial distribution. Finally, we sampled ten text sequences from the policy, and selected the one with the lowest perplexity according to the external language model, and refer to it as \textit{lm-ranking}. 

\section{Results}

\begin{table*}[t!]
\setlength\extrarowheight{-0.5pt}  
\centering
\scriptsize
\resizebox{\textwidth}{!}{
\begin{tabular}{p{0.18\textwidth}|p{0.05\textwidth}p{0.05\textwidth}|p{0.05\textwidth}p{0.05\textwidth}p{0.10\textwidth}p{0.07\textwidth}p{0.07\textwidth}|p{0.07\textwidth}p{0.05\textwidth}}
Method & Score & R@5 & BLEU & Meteor & CIDEr
& ppl-t ($\downarrow$) & ppl-e ($\downarrow$) & sBLEU ($\downarrow$) & peak.($\downarrow$)\\\toprule
Pretrain  & 0.30 & 0.71 & \underline{0.19} & \underline{0.38} & \underline{0.83} & \underline{3.1} & 31 & 0.44 & 0.96 \\
Pretrain + RL fine-tune & 0.44 & 0.86 & 0.17 & 0.34 & 0.70 & 4.0 & 35 & 0.46 &  0.95 \\
\midrule
\midrule
Scratch & 0.17 & 0.47 & 0.05 & 0.08 & 0.10 & $10^{9}$ & $10^6$ & \underline{\textbf{0.14}}  & \underline{\textbf{0.26}}\\
Scratch + KL-task &  0.14 & 0.38 & 0.15 & 0.30 & 0.53 & 92 & $10^2$ & 0.34  & 0.94 \\
Scratch + KL-ext & 0.17 & 0.44 & 0.14 & 0.27 & 0.43 & $10^4$ & 28 & 0.37 & 0.95 \\
\midrule
\algo(Task-LM) & \textbf{\underline{0.56}} & 0.90 &	\textbf{0.17} &	\textbf{0.32} &	\textbf{0.66} &	\textbf{3.4} &	23 &	0.95	&	1.00  \\ 
\algo(Ext-LM) & 0.48 &	\textbf{\underline{0.93}} &	0.08&	0.18&	0.34\scriptsize{($\pm 0.10$)}&	$10^3$ & \underline{\textbf{3.0}} &	0.95&	1.00 \\
\bottomrule
\end{tabular}}
\caption{\small \textbf{CLEVR metrics} on 5k test episodes while training an agent on 20k Images over 50k train episodes. Scores are averaged over the three decoding procedures mentioned in Section~\ref{subsec:decoding:methods} and over 5 seeds; standard deviations are displayed when greater than $0.01$ for accuracy metrics. We here report the models with the highest task-success:, i.e. the \textit{scratch+KL} baselines with $\lambda_{KL}=0.1$, and the truncation model with a probability threshold, $\ptharg{0.05}$. Overall best values are underlined, best values without task-data (from scratch) are in bold.}
\label{table:CLEVR_maintable}
\vskip -0.5em
\end{table*}

\begin{table*}[t!]
\centering
    \setlength\extrarowheight{-2pt}  
		\centering
        \includegraphics[scale=0.25]{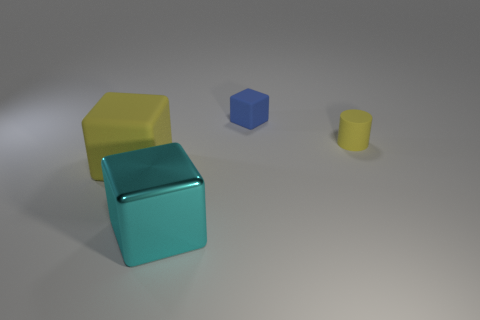}
	\hfill

\vspace{1mm}
		\scriptsize
		\begin{tabular}{p{0.15\textwidth}p{0.72\textwidth}}
			Human           & There is a blue thing that is the same shape as the big cyan metallic object ; what is its size?  \quad \textbf{A:Small} \\
			\midrule
			pretrain        & There is a red metallic object that is the same size as the yellow rubber block ; what is its size? \\
		    pretrain + RL   & What size is the thing that is the same color as the matte cube ? \makebox[0pt][l]{$\square$}\raisebox{.15ex}{\hspace{0.1em}$\checkmark$} \\ \midrule
			scratch         & size sphere small blue or a yellow green large else in cylinders cubes color and how matte objects cube  \\
			scratch+KL-task & How big is the shiny cylinder ? \\
			scratch+KL-ext  & How many other objects in the are of same color as that shiny object ?\\ \midrule
			\algo(Task-LM) & How big is the thing that is to the right of the big matte thing ? \makebox[0pt][l]{$\square$}\raisebox{.15ex}{\hspace{0.1em}$\checkmark$} \\
			\algo(Ext-LM)   & What is the size of the thing that is right of the big cyan thing and is the same shape? \makebox[0pt][l]{$\square$}\raisebox{.15ex}{\hspace{0.1em}$\checkmark$}  \\
			\bottomrule
		\end{tabular}
	\vspace{0.5em}\\
		\centering
        \includegraphics[scale=0.35]{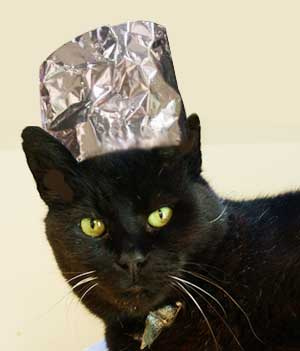}
	\hfill

	\vspace{1mm}
	\small
		\scriptsize
		\begin{tabular}{ll}
			\toprule
			Human            & What color is the cat  \quad \textbf{A:Black} \\
			\midrule
			pretrain        & What color is the cat’s collar? \makebox[0pt][l]{$\square$}\raisebox{.15ex}{\hspace{0.1em}$\checkmark$}  \\
		    pretrain + RL   & What color is the cat? \makebox[0pt][l]{$\square$}\raisebox{.15ex}{\hspace{0.1em}$\checkmark$}  \\ \midrule
			scratch         & AmazingAmazingAmazingAmazingAmazingAmazingAmazing \\
			scratch+KL-task & What color is their hat of the fingers of this? \\
			scratch+KL-ext  & The the first time is a bit of the way \\ \midrule
			\algo(Task-LM)  & What color is her outfit? \makebox[0pt][l]{$\square$}\raisebox{.15ex}{\hspace{0.1em}$\checkmark$} \\
			\algo(Ext-LM)   & What color can these cats look like in real life? \makebox[0pt][l]{$\square$}\raisebox{.15ex}{\hspace{0.1em}$\checkmark$} \\
			\bottomrule
		\end{tabular}
	\captionof{figure}{\small Samples on CLEVR and VQA: the checkbox indicates that the question generates the correct answer.} 
    \label{fig:samples}
    \vskip -1em
\end{table*}

\subsection{CLEVR results}

\paragraph{Quantitative performance:} 
In Table~\ref{table:CLEVR_maintable}, vanilla RL from scratch fails to have a decent performance even with synthetic language. Besides, adding a KL regularisation term does kick-start the learning process. Yet, as soon as we apply the dynamic truncation, \algo matches the pretrained baselines performance when using the external LM, and even outperforms them with the task-specific LM. In this synthetic VQG setting, \algo seems to be a viable and promising procedure to learn a RL language agent from scratch. Pretrained baselines have high language scores when assessed with dataset-based metrics, e.g BLEU or task-perplexity. Yet, they also remain close to the original dataset distribution with a medium external perplexity. Noticeably, \algo with the task-specific LM follows the same pattern. On the other hand, \algo with the external LM reports poor dataset-based language scores, while maintaining a low external perplexity. Therefore, \algo seems to correctly capture the language distribution of the initial LM. As the performance score is high when using an external LM, it suggests that our approach can learn a policy on a language task without the need of a task-related dataset. 
Less positively, \algo diversity metrics suggest potential mode collapse, with a high peakiness and self-BLEU score.

\paragraph{Qualitative performance:}
We display qualitative samples In Figure~\ref{fig:samples} and Appendix~\ref{app:samples}. 
On the one hand, the pretrained baselines generate either a question inconsistent with the visual context, or which fails to answer the expected answer. They inaccurately capture the pragmatics of the task. On the other hand, \algo generate adequate questions, resulting in the expected answer. Interestingly, they are often grounded with different objects of the image. It is remarkable that \algo with a generic LM still manages to capture the necessary subtleties of VQG, without any prior task knowledge. Despite a peaky distribution, \algo has moderate repetitions across images, and is mostly overconfident. As for the scratch+KL samples, they are either not grounded, or showcase degenerated language. 

\begin{table}[ht!]
    \setlength\extrarowheight{-0.5pt}  
\scriptsize
\centering
\begin{tabular}{p{0.08\textwidth}|p{0.04\textwidth}|p{0.04\textwidth}p{0.065\textwidth}p{0.06\textwidth}|p{0.07\textwidth}}
Trunc. & Score & BLEU & CIDEr
& ppl-e($\downarrow$) & sBLEU($\downarrow$) \\\toprule
\multicolumn{3}{l}{\textbf{\algo(Task-LM)}} \\
\hline
$\topk$ & 0.50 & 0.12 & 0.32 & 100 & 0.93 \\
$\pth$ & \underline{\textbf{0.54}} & 0.17 & 0.65 & 24 & 0.95 \\
$\topp$ & 0.51 & 0.17 & 0.69 & \textbf{12} & 0.96 \\
$\sample$ & 0.50 &	\underline{\textbf{0.18}} &	\underline{\textbf{0.73}} &	16 &	\underline{\textbf{0.89}} \\
\toprule
\multicolumn{3}{l}{\textbf{\algo(Ext-LM)}} \\
\hline
$\topk$  & \textbf{0.52} & 0.06 & 0.15 & 151 & 0.94 \\
$\pth$ &  0.48&	0.08&	0.34\scriptsize{($\pm 0.10$)}&	3.0 &	0.95 \\
$\topp$ &  0.45&	0.10&	0.40\scriptsize{($\pm 0.17$)}&3.3 &	\textbf{0.92} \\
$\sample$ & 0.41&	\textbf{0.13}&	\textbf{0.46}\scriptsize{($\pm 0.16$)}&	 \underline{\textbf{2.7}} &	\textbf{0.92} \\
\bottomrule
\end{tabular}
\caption{\small  \textbf{CLEVR task:} Truncation functions with  parameters: $\topkarg{10}$, $\ptharg{0.05}$ $\topparg{0.85}$, $\samplearg{20}$. Overall best values are underlined, best values for each \algo algorithms are in bold.
}
\label{table:abl:CLEVR_trunc_functions}
\vskip -1.1em
\end{table}

\paragraph{Truncation function in CLEVR:}
In Table~\ref{table:abl:CLEVR_trunc_functions}, we evaluate the different truncation functions defined in Section~\ref{sec:truncation:model}. 
While all truncation methods report similar task performance, the dynamic truncation functions, i.e. $\pth$, $\topp$ and $\sample$, outperform the $\topk$ regarding language metrics. Interestingly, the $\sample$ one, which generates a stochastic truncated action space, while having a lower performance, yields to the most correct and diverse language, with higher language scores and a lower self-BLEU. A stochastic action space might be harder to explore efficiently for reaching good task-solving abilities, but might strengthen the agent language generation properties.

\begin{table*}[t!]
\setlength\extrarowheight{-0.5pt}  
\scriptsize
\resizebox{\textwidth}{!}{\begin{tabular}{p{0.22\textwidth}|p{0.05\textwidth}p{0.05\textwidth}|p{0.05\textwidth}p{0.05\textwidth}p{0.05\textwidth}p{0.07\textwidth}p{0.07\textwidth}|p{0.085\textwidth}p{0.05\textwidth}}
Method & Score & R@5 & BLEU & Meteor & CIDEr
& ppl-t ($\downarrow$) & ppl-e ($\downarrow$) & sBLEU ($\downarrow$) & peak.($\downarrow$)\\\toprule
Pretrain  &  0.38 & 0.59 & 0.30 & 0.40 & 0.93 & \underline{12} & \underline{24} & 0.80 & 0.99 \\
Pretrain + RL fine-tune &  \underline{0.41} & \underline{0.63} & \underline{0.31} & \underline{0.41} & \underline{0.98} & 21 & 50 & 0.78 & 0.99 \\
\midrule
\midrule
Scratch & 0.01 & 0.04 & 0.00 & 0.00 & 0.00 & $10^7$ & $10^6$ & 0.75 & 1.00 \\
Scratch + KL-task & 0.11 & 0.29 & \textbf{0.24} & \textbf{0.27} & \textbf{0.24} &  $10^2$ & $10^2$ & 0.27 & 0.74 \\
Scratch + KL-ext & 0.01 & 0.05 & 0.06 & 0.04 & 0.01 & $10^6$ & $10^3$ & \underline{\textbf{0.10}} & \underline{\textbf{0.20}} \\
\midrule
\algo(Task-LM) & \textbf{0.35} & \textbf{0.56} & 0.21 & 0.15 & 0.11 & \textbf{24} & $10^2$ & 0.78 & 0.99 \\
\algo(Ext-LM) & 0.34 & 0.52 & 0.18 & 0.15 &0.04 & $10^2$ & \underline{\textbf{24}} & 0.83 & 0.99 \\
\bottomrule
\end{tabular}}
\caption{\small \textbf{VQAv2} metrics on 20k test episodes with 100k train episodes. Scores are averaged over the three decoding procedures. 
We report the models with the highest task-success, i.e. \textit{scratch+KL} with $\lambda_{KL}=0.05$, and truncation with a probability threshold, $\ptharg{0.005}$ for \algo with (Task-LM) and $\ptharg{0.0075}$ for (Ext-LM). Overall best values are underlined, best values without task-data are in bold.
}
\label{table:VQAv2_maintable}
\end{table*}

\begin{figure*}[t!]
\captionsetup[subfigure]{justification=centering}
\begin{subfigure}{.25\textwidth}
  \centering
  \includegraphics[width=0.98\linewidth]{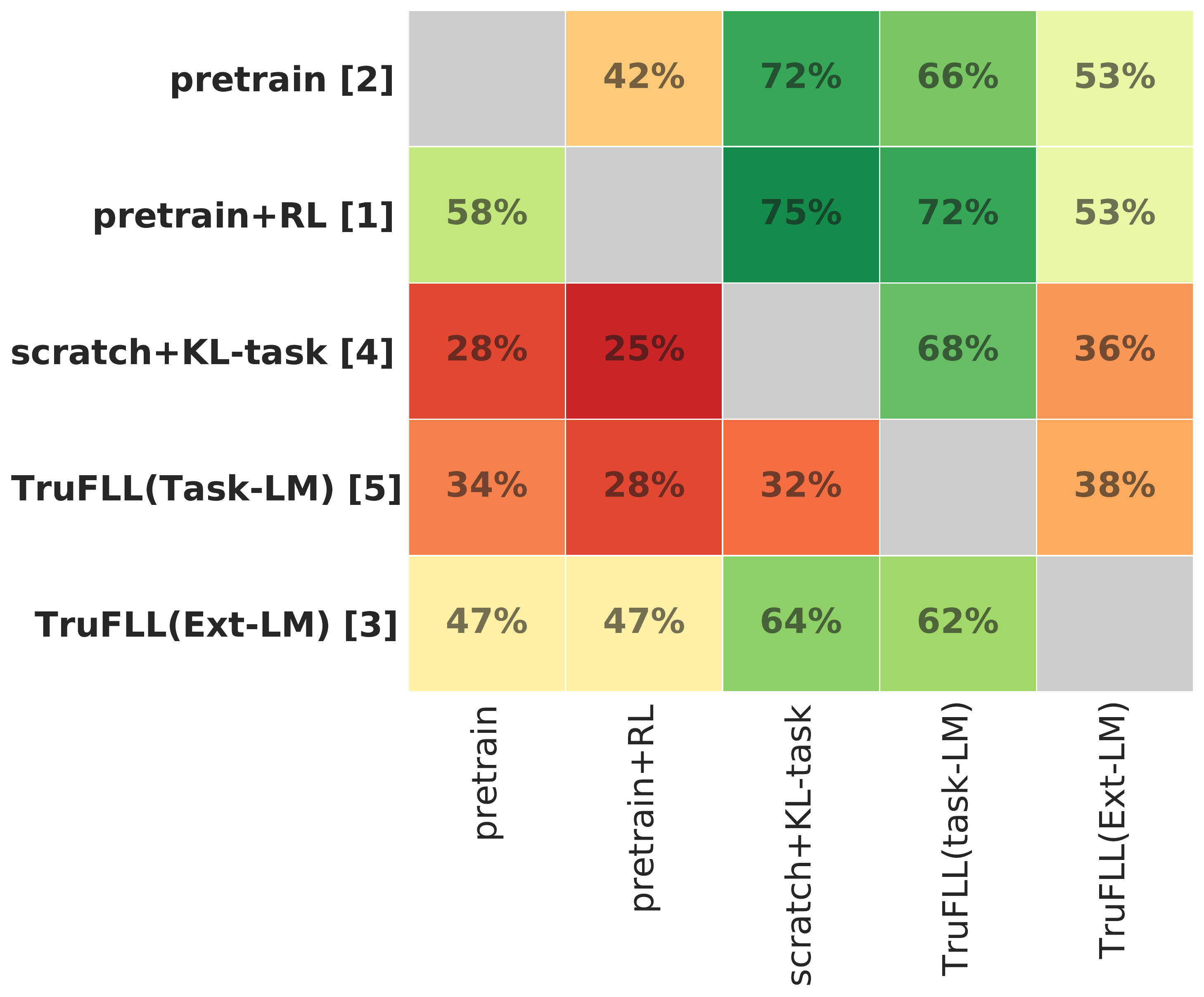}
  \caption{\small \textbf{Language Quality} \\}
  \label{fig:sfig1}
\end{subfigure}%
\begin{subfigure}{.25\textwidth}
  \centering
  \includegraphics[width=0.98\linewidth]{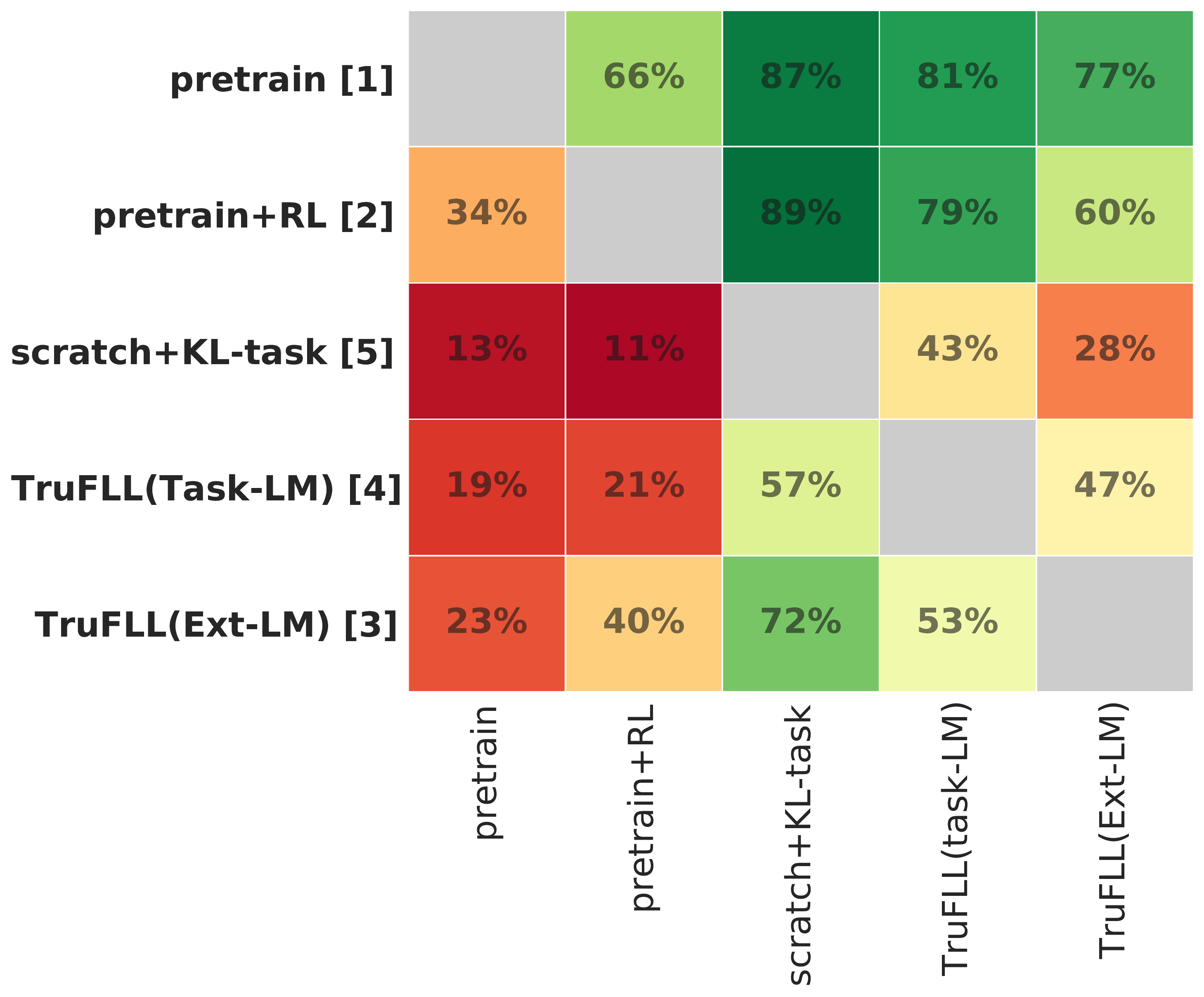}
  \caption{\small \textbf{Language Grounding}}
  \label{fig:sfig2}
\end{subfigure}
\begin{subfigure}{.25\textwidth}
  \centering
  \includegraphics[width=0.98\linewidth]{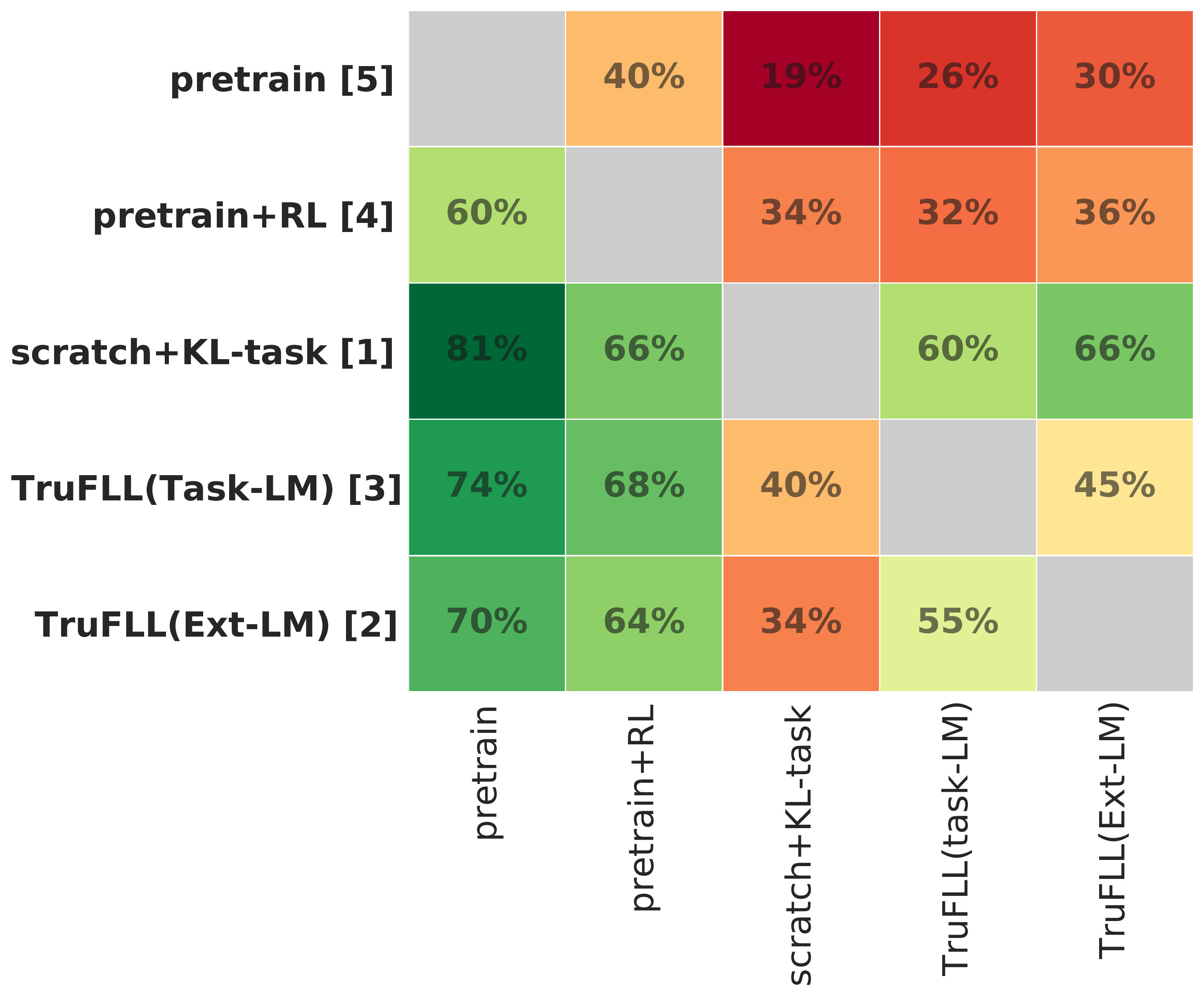}
  \caption{\small \textbf{Diversity/Originality}}  
  \label{fig:sfig3}
\end{subfigure}
\begin{subfigure}{.2\textwidth}
  \centering
  \vskip -0.5em
  \includegraphics[width=1.2\linewidth]{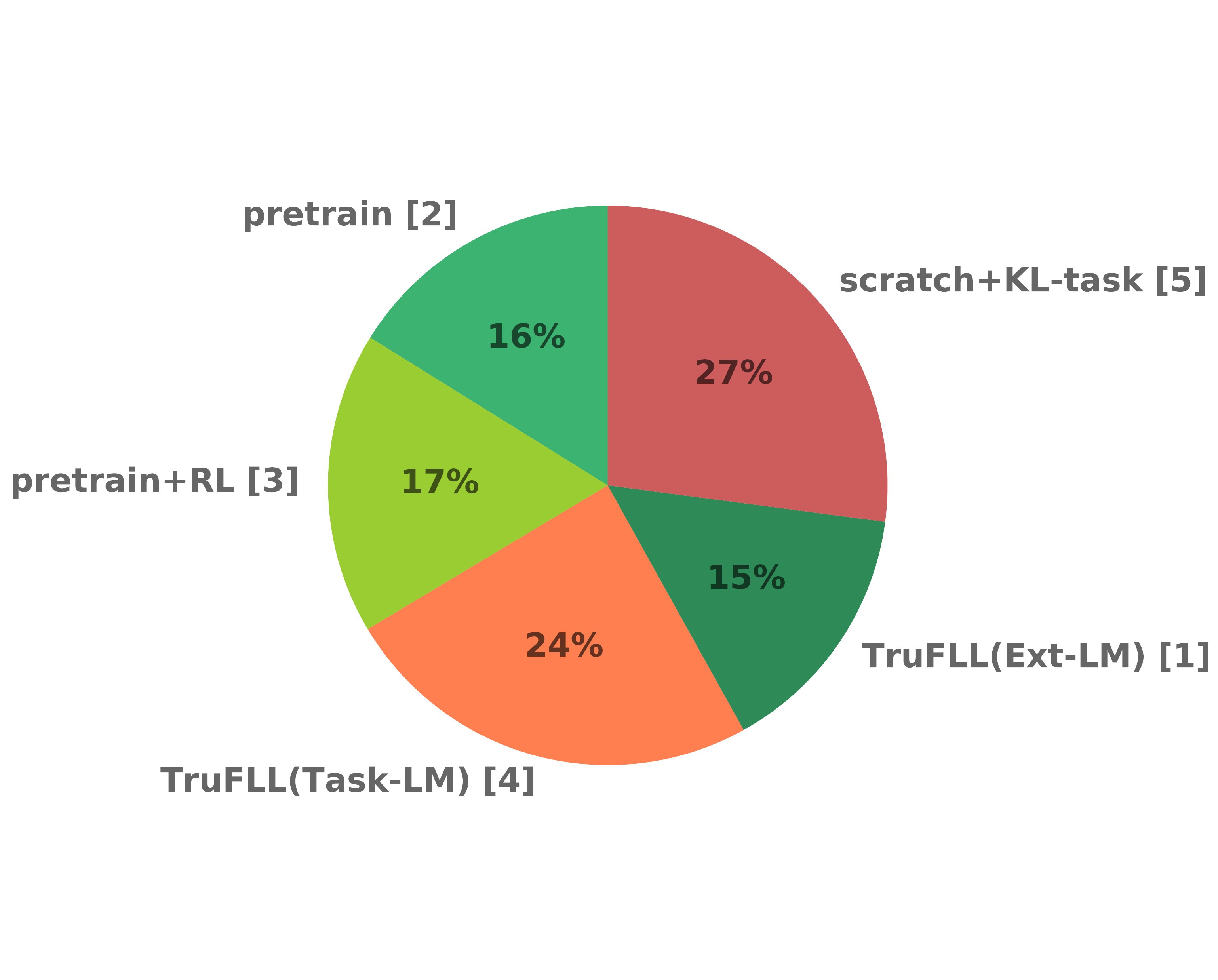}
  \vskip 0.9em
  \caption{\small \textbf{Syntax errors}} 
  \label{fig:sfig4}
\end{subfigure}
\caption{\small \textbf{
VQAv2} results for Human Evaluation study detailed in Section~\ref{subsec:evaluation-metrics}. Matrices (a),(b) and (c) are pairwise comparisons: each cell displays the proportion of questions chosen for the models in the row (bold) when compared to the concurrent model in the column. Figure (d) displays the proportion of incorrect questions coming from each model among all incorrect samples.
In all figures, bracket numbers indicates the model rank per criteria, from 1="best" to 5="worst".}
\label{fig:human_eval_details}
\vskip -0.em
\end{figure*}

\subsection{VQAv2 task}
\label{subsec:exp:VQAv2}

In CLEVR, we observe that \algo seems a promising approach to learn a language policy from scratch by solely interacting with another language system. 
We scale our approach to natural language with large vocabulary (15k tokens) through the VQAv2 dataset.

\paragraph{Quantitative performance:} 
Table~\ref{table:VQAv2_maintable} reports the VQAv2 results, for which \algo and the baselines present a similar trend than on CLEVR. 
First, the scratch baselines keep failing to learn a valuable policy, with performance scores and n-grams metrics close to zero. Although \algo does not outperform the performance of the pretrained baselines anymore, it still leads to similar performances, and satisfactory language scores. The similarity between \algo(Task-LM) and \algo(Ext-LM) results suggests that the truncation approach is viable when using a generic LM whose original vocabulary distribution differs from the task. Interestingly, \algo displays a self-BLEU score similar to the pretrained baselines. This suggests that the poor diversity behavior observed on CLEVR is likely attributable to the small vocabulary and synthetic language distribution. 

\paragraph{Qualitative performance:} 
In Figure~\ref{fig:samples} and Appendix~\ref{app:samples}, we display question samples for all models. \algo and the pretrained baselines successfully generate a question giving the expected answer ("Black"), while the RL from scratch baselines fail, and even showcase degenerated language. 
Pretrained baselines tend to output a question closer to the reference question whereas \algo outputs original questions which differs from the VQA distribution, yet consistent with the context. 

\paragraph{Human Evaluation:}
Figure~\ref{fig:human_eval_details} details the Human Evaluation results. 
Among the RL from scratch baselines, we selected scratch+KL-task as the only model producing sometimes meaningful questions. Yet, it fails to generate correct and grounded language; it is thus not a viable approach despite its diverse output. In line with the automatic metrics, the supervised baselines produce the best language, while being accurately grounded. Yet, they exhibit significantly less diversity with the reference language; this suggests in particular that pretrain+RL fails to go beyond the initial task-data distribution. Finally, unlike \algo(Task-LM) which suffers from syntactic errors, \algo(Ext-LM) produces language that qualitatively competes with pretrain models (53\%), with a similar ratio of syntactic uncorrect samples. Although its questions are less grounded, they are diverse, which suggests that they follow a different distribution from the initial VQA dataset. It confirms that \algo(Ext-LM) could be an alternative approach as it has an excellent trade-off between language quality, diversity, and grounding.

\paragraph{Decoding procedure:}
In Table~\ref{table:abl:VQA_text_gen}, we evaluate the text sampling procedures described in Section~\ref{subsec:decoding:methods}. While greedy decoding produces the best outcome for pretrained models, lm-ranking provides an excellent trade-off between task performance and language quality with RL-based methods. As PG solely optimizes the task success ratio, this may reduce overall language quality, 
the re-ranking thus retrieves the best syntactically sentences a posteriori. 

\begin{table}[h!]
    \setlength\extrarowheight{-0.5pt}  
\centering 
\scriptsize
\begin{tabularx}{0.45\textwidth}{cc|c|ccc}
Method & Text-gen & Score & BLEU & CIDEr
& ppl-e \\\toprule
          & greedy &        0.40 &  \underline{0.32} &  1.01 &         51 \\
pretrain & sampling &        0.37 &  0.30 &  0.88 &         62 \\
         & lm-ranking &        0.37 &  0.14 &  0.87 &         54 \\
\midrule
 & greedy & \underline{0.42} & \underline{0.32} & \underline{1.05} & 55 \\
pretrain + RL & sampling & 0.40 & 0.30 & 0.92 & 71 \\
 & lm-ranking & 0.40 & 0.31 & 0.99 & 26 \\
\midrule
\midrule
            & greedy & \textbf{0.36} & 0.20 & 0.11 & 366 \\
\algo(Task-LM) & sampling & 0.35 & 0.20 & 0.11 & 337 \\
            & lm-ranking & 0.34 & \textbf{0.21} & 0.11 & 95 \\
\midrule
            & greedy & \textbf{0.36} & 0.18 & 0.04 & 25 \\
\algo(Ext-LM) & sampling & 0.34 & 0.18 & 0.04 & 28 \\
            & lm-ranking & 0.33 & 0.19 & \textbf{0.15} & \underline{\textbf{20}} \\

\bottomrule
\end{tabularx}
\caption{\small \textbf{VQAv2:} Ablation on the sampling methods. Overall best values are underlined, \algo best values are in bold.}
\label{table:abl:VQA_text_gen}
\vskip -0.5em
\end{table}

\subsection{Discussion}
\label{sec:discussion}

\paragraph{Removing the truncation at evaluation with off-policy RL.}
So far, \algo directly learns the truncated policy over the truncated vocabulary $\Voc_t^-$ in an on-policy scheme. Hence, the algorithm requires the truncation, and a fortiori the language model, at test time. In this section, we investigate if we can directly learn a policy over the full vocabulary, and thus removing the truncation at test time. In such a  setting, we adopt an off-policy training scheme, where the trajectories used to learn the behavior $\policy$ at training time are sampled under a different policy, the truncated policy $\policy^{-}$.
Thus, we need to unbiased the PG by using an importance sampling term between the exploratory policy $\policy^{-}$ and the behavior policy $\policy$~\cite{degris2012off}. %
Formally, the off-policy PPO loss is defined by:
\begin{equation*}
    L^{off}_{PPO} = \mathbb{E}_{\exploration}\big[\min (\bar{\rho}_t A_t, \clip(1-\epsilon, \bar{\rho}_t, 1 + \epsilon)A_t)\big], 
\label{eq:offpolicyPPO}
\end{equation*}  
where $\bar{\rho} = \frac{\policy(a_t|s_t)}{\policyold(a_t|s_t)} * \frac{\policyold(a_t|s_t)}{\explorationold(a_t|s_t)}$ is the new ratio.\footnote{Note that we did not simplify the expression to highlight the importance sampling ratio.} 

Table~\ref{table:abl:offpolicy_vs_onpolicy} displays the on-policy and off-policy results on both VQG tasks for \algo(task-LM), and is further detailed in Appendix~\ref{app:subsec:discussion}.
We also monitor the probability mass of the policy attributed to the truncated action space ($\mathrm{sumVA}$). 
The policy only samples words within the truncated action space when $\mathrm{sumVA}=1$, without needing the truncation. 
On CLEVR, the $\mathrm{\algo}_{\mathrm{off}}$ has lower - yet close - performance on language and task scores than \algo. As its $\mathrm{sumVA}$ ratios are very close to 1, the agent has learned to generalize over the full vocabulary. However, the approach does not manage to sufficiently scale to VQAv2. It 
could be improved with regularisation techniques and the use of TruFLL within state-of-the-art off-policy RL algorithms. We leave such possibilities to future works.

\begin{table}[h!]

\scriptsize
\centering
\begin{tabular}{p{0.105\textwidth}|p{0.03\textwidth}|p{0.025\textwidth}p{0.025\textwidth}p{0.035\textwidth}|p{0.04\textwidth}p{0.03\textwidth}}
Algo & Score & BLEU & CIDEr & ppl-e & sBLEU & sumVA \\\toprule
\multicolumn{6}{l}{\textbf{CLEVR}} \\
\hline
\algo & 0.56 & 0.17 & 0.06 & $10^3$ & 0.78 & N.A \\
$\mathrm{\algo}_{\mathrm{off}}$ & 0.50 & 0.14 & 0.43 & $10^4$ & 0.88 & 0.96 \\
\toprule
\multicolumn{6}{l}{\textbf{VQAv2}} \\
\hline
\algo & 0.35 & 0.21 & 0.11 & $10^4$ & 0.36 & N.A \\
$\mathrm{\algo}_{\mathrm{off}}$ & 0.07 & 0.03 & 0.01 & $10^4$ &  0.05 & 0.08 \\
\bottomrule
\end{tabular}
\caption{\small On-policy vs. off-policy scores: when training with an off-policy loss, we remove the truncation at test time.}
\label{table:abl:offpolicy_vs_onpolicy}
\vskip -0.5em
\end{table}

\paragraph{Additional experiments.}
We sweep over truncation hyper-parameters in Table~\ref{app:table:abl:CLEVR_trunc_functions_sweep} of Appendix~\ref{sec:app:addtional_xp}. In Table~\ref{app:table:CLEVR_BLEU}, we observe that rewarding an agent with a BLEU score is sub-optimal in both language and task scores on CLEVR. In VQA, we apply temperature scheduling on the LM to perform fine-grained truncations in Table~\ref{app:table:abl:VQA_temp_schedules} of~\ref{sec:app:temperature}. Finally, we explore \algo with a pre-training phase in Table~\ref{table:abl:truffl_pretrain}. 

\section{Related work}
\label{sec:litterature}
\paragraph{Reinforcement Learning and NLP Tasks.}
Following~\citep{singh2002optimizing,lemon2007machine}, recent RL-based task-oriented dialogues~\citep{de2017guesswhat,das2017visual,lewis2017deal, narasimhan2015language} 
have been developed. 
There, the agent is generally pretrained with a SL phase followed by a RL fine-tuning phase to learn the policy's language model. \citet{yang2018visual,fan2018reinforcement} focused on tackling VQG tasks with RL, respectively on CLEVR and on the VQG dataset. 
Yet, the former uses slot filling with template questions, while the later computes a mixed objective with a MLE loss using ground-truth sentences. 
\citet{bahdanau2016actor,rennie2017self} use RL to train language models as an alternative to SL to prevent typical text degeneration issues, but within training algorithms relying on ground-truth examples from labelled datasets. 
\paragraph{RL methods for Language Action Spaces.}  

Several RL algorithms have been developed to tackle  large discrete action spaces. Hence, \citet{dulac2015deep, tennenholtz2019natural, chandak2019learning} embed the actions into a continuous action space, and then use classic RL algorithms to learn a policy over this continuous space. 
\citet{zahavy2018learn,seurin2020im} proposes Q-learning algorithms with an elimination signal to eliminate forbidden actions. 
Closer to our work, a few algorithms \citep{ammanabrolu2018playing} use the structure of language to prune the action space of text-based games,
but within value-based algorithms, which are less scalable to large vocabularies. 
Similarly to \algo, CALM~\citep{yao2020keep} combines a pretrained language model to prune the action space with a Deep-Q network, aka DRNN~\citep{he2016deep}.
Yet, its truncation language model remains fine-tuned on the RL dataset. Besides, CALM is only evaluated on a vocabulary of 697 tokens, and on 4-words action sequences. 

\paragraph{Learning Language Models from scratch.}  
\cite{ziegler2019fine, garg2021unsupervised} finetune pretrained GPT-2 models with RL for language generation tasks without task-related data, only using reward signals.
Yet, they still face optimization and computational challenges~\citep{parisotto2020stabilizing}.

\section{Conclusion}
\label{sec:conclusion}
We proposed \algo, an original approach to learn a natural language generation (NLG) task from scratch using RL. To our knowledge, this is the first RL-based algorithm dedicated to learning a word-based text-generation task, which does not rely on a pre-training phase while scaling to large vocabularies. Although it comes with its limitations, the truncated RL algorithm provided by \algo gets free from labelled data in task-oriented language models, presents interesting language generation properties, and provides a generic and transferable method to learn any NLG problem. 

\bibliographystyle{apalike}
\bibliography{bibliographie}
\appendix

\section{Dataset and training details}

\subsection{Answer filtering}
For each dataset, we remove $yes$ and $no$ question-answer pairs which frequency largely exceeds other answers, to avoid any bias in the question generation process, as usually done in the VQG litterature~\citep{mostafazadeh2016generating}.

\subsection{Dataset split}
For CLEVR (resp. VQAv2), the RL language agent is trained for 50k (resp. 100k) episodes over the first 20k images (resp. all the images) of the training dataset, and is then evaluated on the first 5k (resp. 20k) images of the validation set. Besides, we uniformly sample the answer in the set of reference answers for each image to reduce the bias in the distribution of answers. Finally, questions are limited to 20 (resp. 10) words.

\subsection{Language Agent Networks and Training}
\label{sec:app:hyperparamenters}
For CLEVR (resp. VQAv2),  we used a single-layer LSTM with 64 (resp. 256) units for the policy network. At every time step, the LSTM input is then the concatenation of the word embedding of dimension 32 (resp. 128), the answer embedding of dimension 32 (resp. 128), and the image representation. For CLEVR, the image representation is extracted from a pretrained ResNet50 and projected into a tensor of size (32,7,7) before being flattened. For VQAv2, the image representation is the average of 200 bounding box features of dimension 1048, extracted from a faster R-CNN~\citep{ren2015faster}. 

We optimize the full loss $L = L_{PPO} + \alpha L_{VF} + \beta L_{E}$ with $\alpha=0.5$, $\beta=0.01$ and a PPO clipping ratio $\epsilon=0.02$ (resp. 0.01) for CLEVR (resp. VQAv2). We use Adam optimizer~\citep{kingma2014adam} with a learning rate (lr) of $10^{-3}$ for \algo and the scratch baseline, $10^{-5}$ (resp. $10^{-6}$) for RL algorithms with a pre-training phase on CLEVR (resp. VQAv2), and $5*10^{-4}$ for models including a KL regularization term. We use a batch size (bs) of 128 for all models except the ones with KL regularization, for which we use a batch size of 64. Finally, for the RL from scratch baselines, we perform gradient clipping (gladclip) of $1$ (resp. $5$) for CLEVR and VQAv2.  

Such hyper-parameters were selected, after conducting an extensive hyper-parameter search. The following values were tested:
$\beta \in \{0.01, 0.02, 0.05, 0.1\}$, $\epsilon \in \{0.01, 0.02, 0.05, 0.1, 0.5, 0.9\}$,  lr $\in \{10^{-6}, 10^{-5}, 10^{-4}, 5*10^{-4}, 10^{-3}, 5*10^{-3}, 10^{-2}, 5*10^{-2}\}$,  gradclip $\in \{\mathrm{None}, 1, 5, 10, 100\}$, bs $\in \{32, 64, 128\}$.  

Additionally, we also tested for VQAv2 policy networks with 64, 256 and 1024 units, with respectively 32, 128  and 512 word embedding dimensions. We kept the network size giving the best performances, i.e. policy network of 256 units and 128 word embedding dimension.

\subsection{Reward formula for VQAv2}
\label{sec:app:reward}
In this section, we detail the reward function used for the VQAv2 task. 
$r(w_t, w_{<t}, c) = \indicatrice{\mathrm{rk}(\answer)\leq 10, t=T-1}e^{-\mathrm{rk}(\answer)/2}$, with $\mathrm{rk}(\answer)$ the rank of the ground-truth answer given by the VQA model, when predicting the actual answer from the terminal state $(c,w_{<T})$. Formally, it is defined as: $$\mathrm{rk}(\answer)=\mathrm{rank}(\mathrm{VQA}(c,w_{<T})[\answer])\,,$$ with $\mathrm{VQA}(c,w_{<T})$ the probability distribution given by the VQA model over the set of answers, and $\mathrm{rank}$ the function which ranks the probability of answer $\answer$ within $\mathrm{VQA}(c,w_{<T})$ probability distribution. 

\section{Additional experiments}
\label{sec:app:addtional_xp}

\subsection{CLEVR}
Table~\ref{app:table:abl:CLEVR_trunc_functions_sweep} displays the complete ablation on the truncation functions with parameters sweep. The 'sizeVA' variable indicates the average size of the truncated action space for each truncation function. Table~\ref{app:table:abl:CLEVR_text_gen_sweep} displays the ablation over the three decoding procedures defined in Section~\ref{subsec:decoding:methods}.
Such an ablation presents a similar pattern than VQAv2 results described in section~\ref{subsec:exp:VQAv2}.  

Finally, Table~\ref{app:table:CLEVR_BLEU} reports CLEVR metrics when using the BLEU score as the reward. While on such a task \algo still exhibits promising language scores, the n-grams metrics remain lower than the pretrained baselines. This illustrates that using a language similarity score as a reward signal is much less interesting than a reward based on a task completion score. 

\begin{table}[h!]
\caption{\small \textbf{CLEVR task:} Ablation on the truncation functions with  parameters sweep. Best values are in bold.}
\label{app:table:abl:CLEVR_trunc_functions_sweep}
\centering\hspace*{-0.5cm}
\small
\begin{tabular}{l|c|ccc|cc}
trunc. & Score & BLEU & CIDEr
& ppl-e($\downarrow$) & sBLEU($\downarrow$) & Size VA \\\toprule
\multicolumn{3}{l}{\textbf{\algo(Task-LM)}} \\
\hline
$\topkarg{10)}$ & 0.50 & 0.12 & 0.32 & $10^2$ &0.93 & 10\\
$\topkarg{20}$ & 0.45& 0.10&0.24&$10^3$&0.87& 20 \\
$\ptharg{0.05}$ & \textbf{0.55}&\textbf{0.18}&0.63& 25 & 0.96 &4.4\\
$\ptharg{0.1}$  & 0.47 & 0.18 & \textbf{0.87} & 6.7 & 0.98 &2.4\\
$\ptharg{1/V}$ & 0.50 & 0.16 & 0.49 & 41 & 0.97 &6.6 \\
$\topparg{0.85}$ & 0.52 &	0.17 & 0.69 & 10.4 & 0.96&4.6 \\
$\topparg{0.9}$ & 0.51&	0.17&0.69&11.5 & 0.96 &5.1\\
$\samplearg{20}$ & 0.50&	0.18&	0.73&	18.9 &	\textbf{0.86}&5.4 \\
$\samplearg{30}$ & 0.50 & 0.18 & 0.73 & 16.1 & 0.89&6.1 \\
\toprule
\multicolumn{3}{l}{\textbf{\algo(Ext-LM)}} \\
\hline
$\topkarg{10)}$  & 0.52 & 0.06 & 0.15 & $10^2$ & 0.94 & 10\\
$\topkarg{20}$  & 0.48 & 0.05 & 0.12 & $10^2$ & 0.89 &20\\
$\ptharg{0.05}$ &  0.48&	0.08&	0.34&	3.03&	0.95 &3.3 \\
$\ptharg{0.1}$ &  0.45&	0.17&	0.74&	\textbf{2.2} &	0.99 &2.1 \\
$\ptharg{1/V}$ &  0.44&	0.11&	0.37&	3.7 &	0.96 & 5.7\\
$\topparg{0.85}$ & 0.45& 0.10 & 0.39 & 3.2 &	0.92 & 4.1 \\
$\topparg{0.9}$ & 0.48 & 0.15 & 0.57 & 2.8 & 0.97 & 4.3\\
$\samplearg{20}$ & 0.45&	0.14&	0.50&	2.4 &	0.92 & 4.1 \\
$\samplearg{30}$ & 0.43&	0.13&	0.46&	2.7 &	0.92 & 4.6 \\
\bottomrule
\end{tabular}
\end{table}

\begin{table}[htb]
\caption{\small \textbf{CLEVR task:} Ablation on sampling methods. Best overall values are underlined, while best values for TruFLL are in bold.
}
\label{app:table:abl:CLEVR_text_gen_sweep}
\centering 
\small
\begin{tabularx}{0.6\textwidth}{cc|c|ccc}
method & text-gen & score & BLEU & CIDEr
& ppl-e \\\toprule
          & greedy &  0.32 &  \underline{0.22} & \underline{1.01} & 14 \\
pretrain & sampling &  0.29 &  0.17 &  0.76 & 58 \\
         & lm-ranking & 0.28 &  0.18 &  0.73 & 20 \\
\midrule
 & greedy & 0.53 &  0.18 &  0.73 & 24 \\
pretrain + RL & sampling & 0.40 &  0.16 &  0.68 &  39 \\
 & lm-ranking & 0.40 & 0.17 &  0.68 &  5 \\
\midrule
                 & greedy & \underline{\textbf{0.57}}&0.17&0.65& 39  \\
Task-LM & sampling & 0.55&\textbf{0.17}&\textbf{0.66}& 24 \\
                 & lm-ranking &0.51&0.16&0.65& 9 \\
\midrule
            & greedy & 0.48&	0.09&	0.34\scriptsize{($\pm 0.11$)}&	3.0	 \\
Ext-LM & sampling & 0.48&	0.10 &	0.35\scriptsize{($\pm 0.11$)}&	3.1 \\
            & lm-ranking & 0.48&	0.06&	0.34\scriptsize{($\pm 0.11$)}&	\underline{\textbf{2.9}} \\
\bottomrule
\end{tabularx}
\end{table}

\begin{table*}[h!t]
\caption{\small \textbf{CLEVR, BLEU reward}. Scores are averaged over the three decoding procedures detailed in Section~\ref{subsec:decoding:methods} and over 5 seeds, standard deviation are displayed whenever greater than $0.01$ for accuracy metrics. We here report the models with the highest task-success, i.e. the \textit{scratch with KL regularization} baseline with $\lambda_{KL}=0.1$, and the truncation model with a probability threshold, $\ptharg{0.05}$. Baseline and Metrics are respectively detailed in Section~\ref{subsec:evaluation-metrics} and~\ref{subsec:baseline}. Best overall values are underlined, while best values for models without task-data (i.e RL from scratch algorithms) are in bold.}
\label{app:table:CLEVR_BLEU}
\centering\hspace*{-0.6cm}
\small
\resizebox{\textwidth}{!}{\begin{tabular}{p{0.20\textwidth}|p{0.05\textwidth}p{0.08\textwidth}|p{0.05\textwidth}p{0.05\textwidth}p{0.10\textwidth}p{0.07\textwidth}p{0.07\textwidth}|p{0.09\textwidth}p{0.05\textwidth}}
Method & Score & R@5 & BLEU & Meteor & CIDEr
& ppl-t ($\downarrow$) & ppl-e ($\downarrow$) & sBLEU ($\downarrow$) & peak.($\downarrow$)\\\toprule
pretrain & 0.30 & 0.71 & 0.19 & 0.38 & 0.83 & 3.1 & 31 & 0.44 & 0.96 \\
pretrain + RL fine-tune &  \underline{0.34} &  \underline{0.80} &  \underline{0.20} &  \underline{0.38} &  \underline{0.83} &  3.8 & 12 &  0.56 &  0.96 \\
\midrule
scratch & 0.03 &  0.19 &  0.06 &   0.09 &  0.09 & $10^8$ & $10^6$ &  \underline{\textbf{0.13}} &  \underline{\textbf{0.14}}  \\
scratch + KL-task &  0.09 &   0.33\scriptsize{($\pm 0.15$}) &   0.15 &  0.31 &  0.58\scriptsize{($\pm 0.23$}) &  3.8 &  63 &  0.34 &  0.95 \\
scratch + KL-ext & 0.06 &  0.30\scriptsize{($\pm 0.23$}) &  0.13 &  0.25 &  0.42 & $10^3$ & 3.6 &  0.37 & 0.96\\
\midrule
scratch + Truncation-task & \textbf{0.17} &  \textbf{0.51} &   \textbf{0.18} &   \textbf{0.37} &  \textbf{0.80} &   \underline{\textbf{2.6}} & 17 &  0.63 &  1.0 \\
scratch + Truncation-ext & 0.07 & 0.36 &  0.16 &  0.29 &  0.49 &  $10^2$ &  \underline{\textbf{2.3}} &  0.60 & 1.0 \\
\bottomrule
\end{tabular}}
\end{table*}

\subsection{VQAv2}

\paragraph{Temperature scheduling:}
\label{sec:app:temperature}
On the CLEVR task, we observed that dynamic truncations outperform static ones such as $\topk$: indeed, they better take into account the inherent variability of the language structure at the sentence-level. When scaling up to the 15k words of the VQAv2 task, we also dynamically decrease the truncation size through training, by applying a decreasing temperature schedule on the language model.
While temperature scaling \cite{bahdanau2014neural} is usually used at test time to control the smoothness of the language model distribution, temperature schedules during training of language models have been used in several settings \cite{jang2016categorical,zhang2018heated, wang2020contextual}. Formally, $f_{LM}(w_i|w_{<t})$ distribution is computed as $\mathrm{softmax}(x_i)=e^{-x_i/\tau}/\sum_{j} e^{-x_{j}/\tau}$, with $x_j$ the LM logits and $\tau$ the temperature, which decreases from $\tau_{max}$ to $\tau_{min}$ by a factor $T_{F}$ every $T_{u}$ training step.  
In Table~\ref{app:table:abl:VQA_temp_schedules}, both \algo(Task-LM) and \algo(Ext-LM) benefit slightly from truncation with a temperature schedule compared to a vanilla truncation. 
The former displays the best performance/language scores trade-off for the schedule "$\tau$: 3 > 1. \& $T_{u}$=5,000", while the latter has the best metrics trade-off for "$\tau$: 1.5 > 1. \& $T_{u}$=5,000".

\begin{figure}[h]
    \centering
    \includegraphics[width=\textwidth]{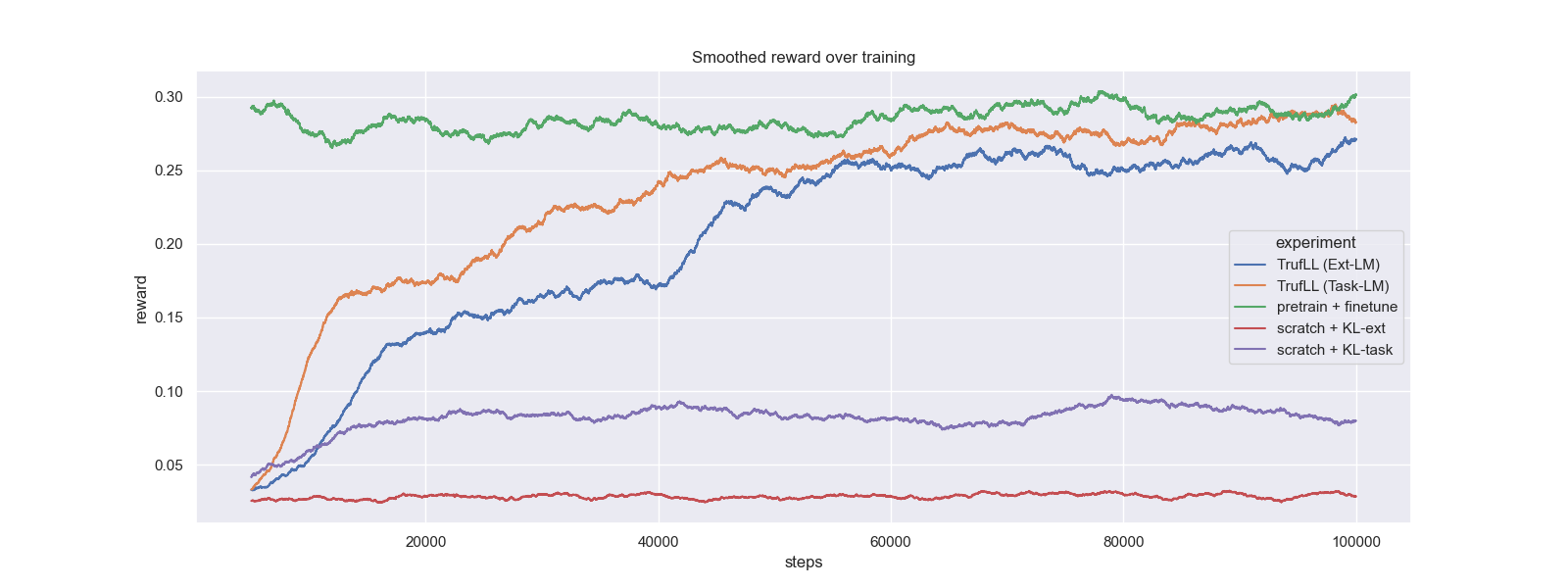}
    \caption{\textbf{VQAv2:} Training curves. Reward is a rolling average over 5000 timesteps.}
    \label{fig:vqa_training_1}
\end{figure}

Finally, Figure~\ref{fig:vqa_training_1} displays the evolution of the training return for \algo and the baselines. As expected, the pretrain+RL fine-tune baseline return does not evolve much, confirming that the policy distribution almost does not shift through the fine-tuning phase. The training curves of \algo present a steady increase in the return until reaching convergence, confirming that our approach, by guiding the exploration of the action space, provides a sufficient learning signal. On the other hand, the scratch+KL baselines stay stuck to a low training return. This suggests that the KL regularization term, while encouraging the policy distribution to resemble the language model distribution, fails to capture the task pragmatics, which requires generating a language that is visually grounded. 

\begin{table}[h!]
\caption{\small \textbf{VQA task:} Ablation on the temperature schedules. "no temp. sch" is a classic truncation without temperature scheduling. We then report different schedules $\tau:\tau_{max}>\tau_{min}$, $T_{u}$, with  $\tau_{max}$, $\tau_{min}$, $T_{u}$, and $T_{f}=0.75$ as defined in section~\ref{sec:app:temperature}. Best values are in bold.
}.
\label{app:table:abl:VQA_temp_schedules}
\centering\hspace*{-0.8cm}
\small
\begin{tabular}{p{0.10\textwidth}p{0.10\textwidth}|p{0.07\textwidth}|p{0.07\textwidth}p{0.07\textwidth}p{0.07\textwidth}|p{0.07\textwidth}}
\multicolumn{2}{l}{Scheduling}& Score & BLEU & CIDEr
& ppl-e($\downarrow$) & sBLEU($\downarrow$) \\\toprule
\multicolumn{4}{l}{\textbf{\algo(Task-LM)}} \\
\hline
\multicolumn{2}{l|}{no temp. sch} & \textbf{0.35} & 0.20 & 0.11 & $10^2$ & 0.78 \\
$\tau$: 1.5 > 1. &$T_{u}$=5,000  & 0.34 &  0.18 & 0.11 & $10^2$ & 0.79 \\
$\tau$: 3 > 1. & $T_{u}$=5,000 & \textbf{0.35} & 0.22 & 0.13 & $10^2$ & 0.76 \\
$\tau$: 1.5 > 1. & $T_{u}$=15,000 & 0.31 & \textbf{0.23} &  \textbf{0.23} & $10^2$ & \textbf{0.73} \\
\toprule
\multicolumn{4}{l}{\textbf{\algo(Ext-LM)}} \\
\hline
\multicolumn{2}{l|}{no temp. sch} & 0.34 & 0.18 & 0.04 & 25 & 0.83 \\
$\tau$: 1.5 > 1. &$T_{u}$=5,000 & 0.33 & 0.19 &  0.05 & \textbf{20} & 0.83  \\
$\tau$: 3 > 1. & $T_{u}$=5,000 & 0.32 &  0.15 &  0.05 &  35 & 0.82 \\
$\tau$: 1.5 > 1. & $T_{u}$=15,000 & 0.29 &  0.16 &  0.08 & 38 & 0.68 \\
\bottomrule
\end{tabular}
\end{table}

\subsection{Additional discussion}
\label{app:subsec:discussion}
\paragraph{\algo with a pre-training phase.}
\label{sec:app:pretrain}
Although \algo aims at providing a robust method to learn a language model (almost) from scratch, we investigate whether such algorithm can be complementary to RL algorithms with a pre-training phase. Therefore, when using the task-related dataset, we evaluate \algo from a pretrained policy, and we refer to it as $\mathrm{\algo}_{\mathrm{pretrain}}$.  

In table~\ref{table:abl:truffl_pretrain}, while on CLEVR, $\mathrm{\algo}_{\mathrm{pretrain}}$ marginally improves the results of the pretrain+RL fine-tune baseline, the combination of \algo with a pre-training phase leads to performance degradation on VQAv2. This suggests that on a large vocabulary task, the language distribution learned by the SL pretrained policy is significantly different from the one learned with \algo. 

\begin{table}[h!]
\caption{\small $\mathrm{\algo}_{\mathrm{pretrain}}$ results on the 2 tasks. Additionally, we report the results for the pretrain+RL fine-tune baseline as a comparison. Best values are in bold.}
\label{table:abl:truffl_pretrain}
\small
\centering
\begin{tabular}{p{0.15\textwidth}|p{0.07\textwidth}|p{0.07\textwidth}p{0.04\textwidth}p{0.07\textwidth}|p{0.07\textwidth}}
Algo & Score & BLEU & CIDEr & ppl-e & sBLEU \\\toprule
\multicolumn{6}{l}{\textbf{CLEVR}}\\
\hline
pretrain+RL & 0.44 & 0.17 & 0.70 & 35 & \textbf{0.46} \\
$\mathrm{\algo}_{\mathrm{pretrain}}$ & \textbf{0.61} & \textbf{0.18} & \textbf{0.77} & \textbf{22} & 0.84 \\
\toprule
\multicolumn{6}{l}{\textbf{VQAv2}}  \\
\hline
pretrain+RL & \textbf{0.41} & \textbf{0.31} & \textbf{0.98} & 50 & \textbf{0.78} \\
$\mathrm{\algo}_{\mathrm{pretrain}}$ & 0.33 & 0.27 & 0.42 & \textbf{35} & 1.0 \\
\bottomrule
\end{tabular}
\end{table}

\paragraph{On-policy \algo versus off-policy \algo.}
To ease off-policy learning, we propose to add a KL-regularization term in the RL loss~\citep{jaques2017sequence, jaques2019way, wu2019alternating}, and refer to it as $\mathrm{\algo_{\mathrm{off, KL}}}$.
Intuitively, it encourages the policy to stay close to the language model's distribution, with a distribution support attributing negligible probabilities to words outside the truncated action space.

Table~\ref{table:abl:offpolicy_vs_onpolicy_sweep} displays the full results of on-policy versus off-policy scores for \algo(Task-LM) and \algo(Ext-LM) on the two tasks. The full results emphasize the challenges of the approach for the large vocabulary of VQAv2. Indeed, on the off-policy setting for such a task, the exploding values for e-ppl suggest that the optimized language agent samples incoherent words taken outside the truncated action space, as corroborated by the low values of the $\mathrm{sumVA}$ ratio.  

Interestingly, while on CLEVR, $\mathrm{\algo}_{\mathrm{off,KL}}$ trades off task performance for language quality when compared to $\mathrm{\algo}_{\mathrm{off}}$, on VQAv2, it mainly provides a better learning signal for the complete (large) vocabulary. In such a setting, it hence improves the global scores of the off-policy version of \algo, and enables a much better generalization at test time of the global policy over the full vocabulary. Yet, keeping truncation at test time remains crucial with large vocabulary. Note that for VQAv2, the poor performances of $\mathrm{\algo}_{\mathrm{off,KL}}$ on the external LM is mainly due to numerical instability challenges when using GPT-2 as the target policy of the KL regularization term.

Additionally, on-policy versus off-policy scores split per sampling procedure are displayed in table~\ref{app:table:abl:onpolicy_offpolicy_textgen}: unsurprisingly, greedy decoding for $\mathrm{\algo}_{\mathrm{off}}$ outperforms the two sampling-based methods, that are more penalized by the imperfect generalization of the optimized policy over the full vocabulary. 

\begin{table}[h!]
\caption{\small On-policy vs. off-policy scores for different variants of \algo: when training with an off-policy loss, we remove the truncation at test time.
$\mathrm{\algo}_{\mathrm{off,KL}}$ is evaluated with $\lambda_{KL}=0.05$. Best values are in bold.}
\label{table:abl:offpolicy_vs_onpolicy_sweep}
\small
\centering
\begin{tabular}{p{0.12\textwidth}|p{0.07\textwidth}|p{0.07\textwidth}p{0.03\textwidth}p{0.07\textwidth}|p{0.07\textwidth}p{0.07\textwidth}}
Algo & Score & BLEU & CIDEr & ppl-e & sBLEU & sumVA \\\toprule
\multicolumn{6}{l}{\textbf{CLEVR}}\\
\hline\hline
\multicolumn{4}{l}{\textbf{\algo(Task-LM)}} \\
\midrule
\algo & \textbf{0.56} & \textbf{0.17} & 0.06 & $10^3$ & 0.78 & N.A \\
$\mathrm{\algo}_{\mathrm{off}}$ & 0.50 & 0.14 & 0.43 & $10^4$ & 0.88 & 0.96 \\
$\mathrm{\algo}_{\mathrm{off,KL}}$ &0.39 & 0.17&\textbf{0.71} &69 &\textbf{0.48} & 0.95 \\
\midrule
\multicolumn{4}{l}{\textbf{\algo(Ext-LM)}} \\
\midrule
\algo & 0.48 & 0.08 & 0.34 & \textbf{3.03} & 0.95 & N.A \\
$\mathrm{\algo}_{\mathrm{off}}$ & 0.41 & 0.10 & 0.35 & $10^5$ & 0.88 & 0.95 \\
$\mathrm{\algo}_{\mathrm{off,KL}}$ & 0.35 & 0.15 & 0.60 & 20 & 0.55 & 0.96 \\ 
\toprule 
\multicolumn{6}{l}{\textbf{VQAv2}}  \\
\hline\hline
\multicolumn{4}{l}{\textbf{\algo(Task-LM)}} \\
\midrule
\algo & \textbf{0.35} & 0.21 & 0.11 & $10^4$ & 0.36 & N.A \\
$\mathrm{\algo}_{\mathrm{off}}$ & 0.07 & 0.03 & 0.01 & $10^4$ &  0.05 & 0.08 \\
$\mathrm{\algo}_{\mathrm{off,KL}}$ &  0.12 &  \textbf{0.24} &  \textbf{0.25} & 10³ &  0.26 & 0.71 \\
\midrule
\multicolumn{4}{l}{\textbf{\algo(Ext-LM)}} \\
\midrule
\algo & 0.34 & 0.18 & 0.04 & \textbf{24} & 0.83 & N.A \\
$\mathrm{\algo}_{\mathrm{off}}$ & 0.09 & 0.04 & 0.01 & $10^4$ & \textbf{0.05} & 0.07 \\
$\mathrm{\algo}_{\mathrm{off,KL}}$ & 0.0 & 0.15 &  0.02 & $10^3$ &  0.19 &  0.47 \\
\bottomrule
\end{tabular}
\end{table}

\begin{table}[h!]
\caption{\small On-policy vs. off-policy scores per decoding procedure: when training with an off-policy loss, we remove the truncation at test time. $\mathrm{\algo}_{\mathrm{off,KL}}$ is evaluated with $\lambda_{KL}=0.05$. Best values are in bold.}
\label{app:table:abl:onpolicy_offpolicy_textgen}
\centering 
\small
\begin{tabularx}{0.6\textwidth}{cc|c|ccc}
method & text-gen & score & BLEU & CIDEr
& e-ppl \\\toprule
\multicolumn{6}{l}{\textbf{CLEVR}} \\
\hline
\hline
\multicolumn{4}{l}{\textbf{\algo(Task-LM)}} \\
\hline
& greedy & \textbf{0.57} & 0.17 & 0.65 & 39 \\
$\mathrm{\algo}$ & sampling & 0.55 & 0.17 & 0.66 & 24 \\
              & lm-ranking & 0.51 & 0.16 & 0.65 & 8.8 \\
\hline
\hline
& greedy & 0.52 & 0.17 &  0.58 & 71 \\
$\mathrm{\algo}_{\mathrm{off}}$ & sampling & 0.49 & 0.16 & 0.59 & $10^5$ \\
                 & lm-ranking & 0.48 & 0.17 & 0.58 & 19 \\
\hline
& greedy & 0.56 & \textbf{0.18} & \textbf{0.78} & 24 \\
$\mathrm{\algo}_{\mathrm{off,KL}}$ & sampling & 0.31 & 0.16 & 0.62 &$10^2$ \\
                 & lm-ranking & 0.31 & 0.18 & 0.74 & 5.8 \\
\midrule
\multicolumn{4}{l}{\textbf{\algo(Ext-LM)}} \\
\midrule
 & greedy & 0.48 &	0.09 &	0.34 & 3.1 \\
\algo & sampling & 0.48&	0.10 &	0.35 &	3.1 \\
            & lm-ranking & 0.48 & 0.06 & 0.34 & 2.9 \\
\hline
\hline
& greedy & 0.42 &  0.10 &  0.38 &  4.4 \\
$\mathrm{\algo}_{\mathrm{off}}$ & sampling & 0.40 & 0.10 & 0.35 & $10^6$ \\
            & lm-ranking & 0.40 & 0.10 & 0.34 & 15 \\
\hline
& greedy & 0.48 & 0.16 & 0.70 & 2.1 \\
$\mathrm{\algo}_{\mathrm{off,KL}}$ & sampling & 0.27 & 0.13 & 0.48 & 55 \\
                 & lm-ranking &0.30 & 0.16 & 0.61 & \textbf{2.0} \\
\toprule
\multicolumn{6}{l}{\textbf{VQAv2}} \\
\hline
\hline
\multicolumn{4}{l}{\textbf{\algo(Task-LM)}} \\
\midrule
 & greedy & 0.36 & 0.20 & 0.11 & 366 \\
\algo & sampling & 0.35 & 0.20 & 0.11 & 337 \\
             & lm-ranking & 0.34 & 0.21 & 0.11 & 95 \\
\hline
\hline
 & greedy & 0.09 & 0.04 & 0.02 & $10^3$ \\
$\mathrm{\algo}_{\mathrm{off}}$  & sampling  & 0.05 & 0.03 & 0.01 & $10^6$ \\
                 & lm-ranking & 0.06 & 0.03 & 0.01 & $10^4$ \\
\hline
& greedy &  0.16 & 0.29 &  0.46 &  38 \\
$\mathrm{\algo}_{\mathrm{off,KL}}$ & sampling & 0.08 &  0.19 & 0.09 & $10^4$ \\
                 & lm-ranking & 0.12 & 0.24 &  0.22 &  $10^2$ \\
\midrule
\multicolumn{4}{l}{\textbf{\algo(Ext-LM)}} \\
\hline
& greedy & \textbf{0.48} & 0.09& 0.34&	3.1 \\
\algo & sampling & 0.48 &	0.10 & \textbf{0.35} & 3.1 \\
            & lm-ranking & 0.48 & 0.06 & 0.34 &	\textbf{2.9} \\
\hline
\hline
& greedy & 0.11 & 0.05 & 0.01 & $10^2$ \\
$\mathrm{\algo}_{\mathrm{off}}$ & sampling  & 0.07 & 0.03 & 0.01 & $10^5$ \\
            & lm-ranking & 0.08 & 0.04 & 0.01 & $10^4$ \\
\hline
& greedy & 0.00 &  \textbf{0.18}  & 0.05 &  27 \\
$\mathrm{\algo}_{\mathrm{off,KL}}$ & sampling & 0.00 & 0.13 &  0.01 &  $10^3$ \\
                 & lm-ranking & 0.00 & 0.16 & 0.02 & $10^2$ \\
\bottomrule
\end{tabularx}
\end{table}

\FloatBarrier
\section{Human Evaluation details}
\label{app:human:eval}
For the Human Evaluation study, we designed one form per participant, with three sections evaluating respectively the language quality, language grounding and diversity criteria. Given the five evaluated models, there are ten different model pairs: each section of the form contains 10 pairwise comparison covering all the possible model pairs for the criteria. Each pairwise comparison is sampled uniformly over the 50 first question samples generated by the algorithms at test time. The evaluation of syntax errors was made within the diversity section: for each questions pair, we asked participants to tick the questions if they are grammatically incorrect. Figure~\ref{app:fig:human:eval:examples} displays one pairwise comparison example for the three sections, and a full form example is available at the following url: \url{https://forms.gle/kkL38x31wF7A9YKx5}. 

\begin{figure}[h!]
\begin{subfigure}{\textwidth}
\centering
\includegraphics[clip,width=0.8\textwidth,height=6cm]{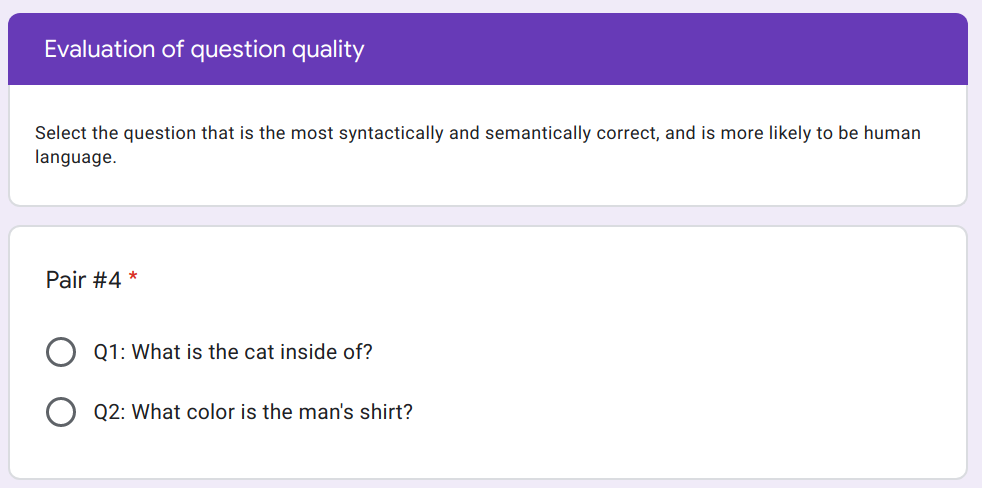}
\caption{Language Quality pairwise comparison}
\end{subfigure}
\begin{subfigure}{\textwidth}
\centering
\includegraphics[clip,width=0.6\textwidth,height=10cm]{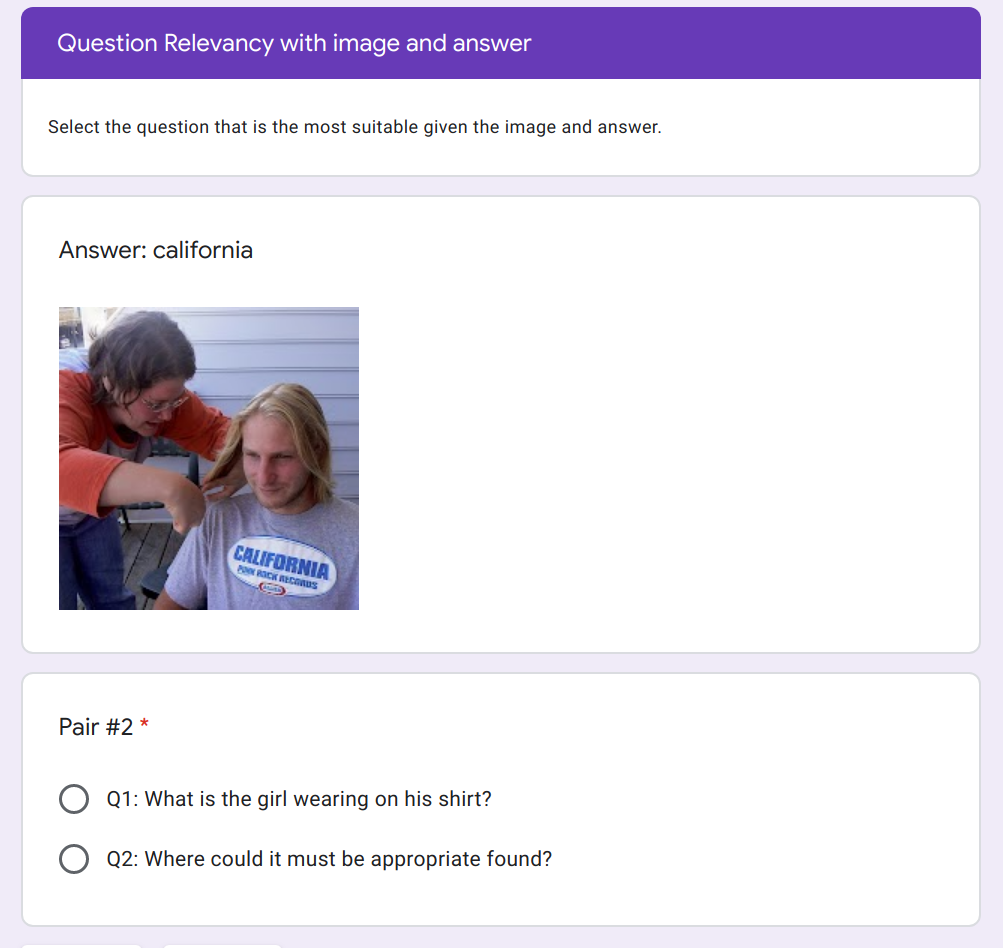}
\caption{Language Grounding pairwise comparison}
\end{subfigure}
\caption{Examples of pairwise comparison for each evaluated criteria.}
\label{app:fig:human:eval:examples}
\end{figure}
\begin{figure}\ContinuedFloat
    \begin{subfigure}{\textwidth}
\centering
\includegraphics[clip,width=0.8\textwidth,height=8cm]{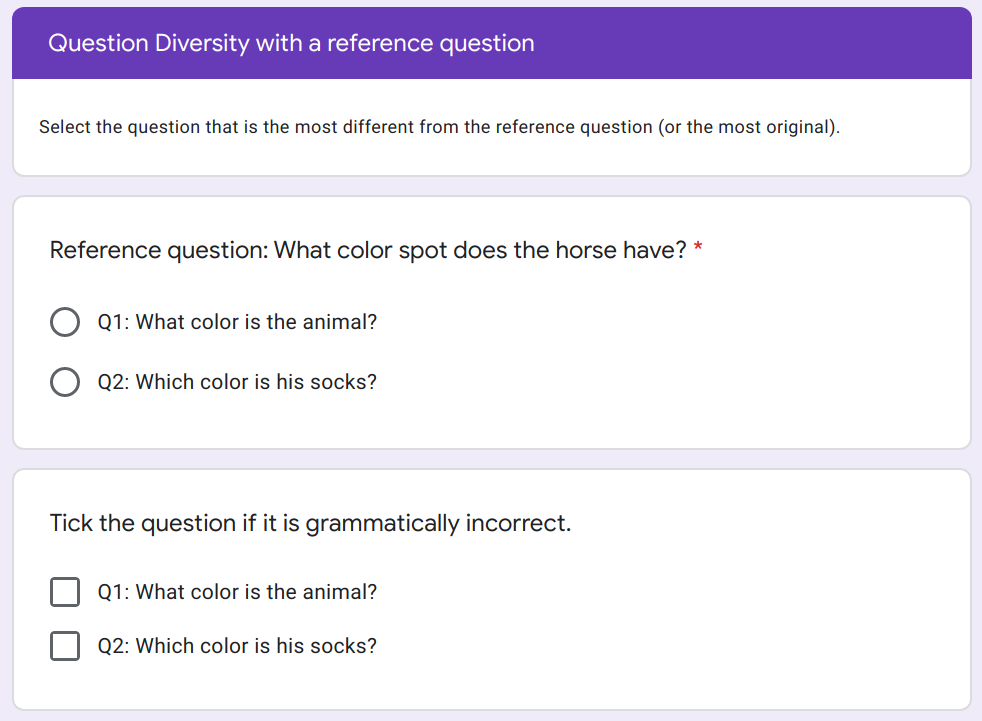}
\caption{Diversity/Originality with reference question. Pairwise comparison and evaluation of syntax errors.}
\end{subfigure}
    \caption{Examples of pairwise comparison for each evaluated criteria. (cont.)}
\end{figure}

\FloatBarrier
\section{Additional VQG Samples}
\label{app:samples}

Figure~\ref{fig:clevrsamples1} and Figure~\ref{fig:clevrsamples2} display the 10 first dialog samples produced at test time on CLEVR, while figures~\ref{fig:vqasamples1}, \ref{fig:vqasamples2}, and \ref{fig:vqasamples3} display the 15 first dialog samples produced at test time on VQAv2.

\begin{table*}[h!]
  \begin{minipage}{0.2\linewidth}
        \includegraphics[width=120px]{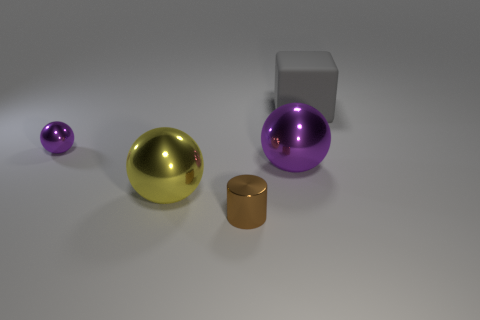}
	\end{minipage}
           \hspace{1.2cm}
 \begin{minipage}{0.8\linewidth}
		\scriptsize
		\begin{tabular}{ll}
			Human           & the big yellow object is what shape ?  \quad \textbf{A:sphere} \\
			\midrule
			pretrain        & there is a small thing that is the same color as the small rubber cylinder ; what is its shape? \\
		    pretrain + RL   & What is on the person's head? \\ \midrule
			scratch         & yellow on or an material ?  \\
			scratch+KL-task & what number of other things are the same shape as the small gray thing ? \\
			scratch+KL-ext  & does that tiny object have objects to its left ?\\ \midrule
			\algo(Task-LM) & what shape is the big thing that is to the right of the big matte thing ? \\
			\algo(Ext-LM)   & what is the shape of the big object that is behind the big yellow thing and is the same color? \\
			\bottomrule
		\end{tabular}
	\end{minipage}
	\vspace{1mm}

  \begin{minipage}{0.2\linewidth}
        \includegraphics[width=120px]{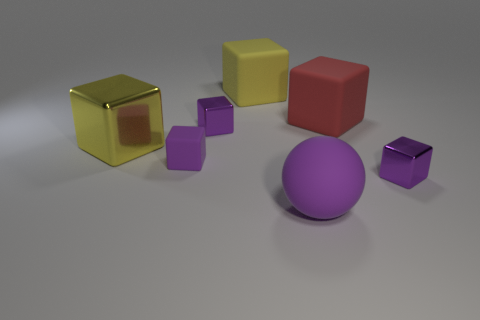}
	\end{minipage}
	 \hspace{1.2cm}
	\begin{minipage}{0.80\linewidth}
		\scriptsize
		\begin{tabular}{ll}
			Human           & what number of other objects are the same size as the red rubber thing ?  \quad \textbf{A:3} \\
			\midrule
			pretrain        & how many other things are there of the same size as the purple rubber thing ? \\
		    pretrain + RL   & what number of objects are either large green metallic objects or tiny objects that are behind the tiny purple metallic thing? \\ \midrule
			scratch         & many tiny cubes things things things things things both things as things that  \\
			scratch+KL-task & what number of other objects are there of the same material as the tiny cyan thing ? \\
			scratch+KL-ext  & are there any blue objects ?\\ \midrule
			\algo(Task-LM) & what number of objects are either big objects in front of the small yellow object or big matte objects? \\
			\algo(Ext-LM)   & how many objects in front of the big object ? \\
			\bottomrule
		\end{tabular}
	\end{minipage}
	\vspace{1mm}	

  \begin{minipage}{0.2\linewidth}
        \includegraphics[width=120px]{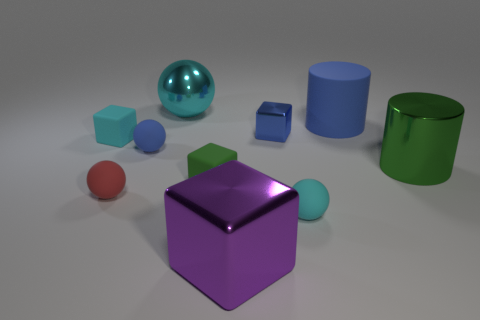}
	\end{minipage}
	 \hspace{1.2cm}
	\begin{minipage}{0.80\linewidth}
		\scriptsize
		\begin{tabular}{ll}
			Human           & what number of other things are there of the same material as the large green object?  \quad \textbf{A:3} \\
			\midrule
			pretrain        & how many other things are there of the same size as the purple rubber cylinder ? \\
		    pretrain + RL   & what number of objects are either tiny cyan things or big cyan things ? \\ \midrule
			scratch         & many tiny cubes things things things things things both things as things that  \\
			scratch+KL-task & what number of other objects are the same shape as the small yellow object ? \\
			scratch+KL-ext  & how many things does that large thing have to its behind ?\\ \midrule
			\algo(Task-LM) & what number of other things are there of the same size as the green cylinder ? \\
			\algo(Ext-LM)   & how many objects in front of the in the cylinder ? \\
			\bottomrule
		\end{tabular}
	\end{minipage}
	\vspace{1mm}	

  \begin{minipage}{0.2\linewidth}
        \includegraphics[width=120px]{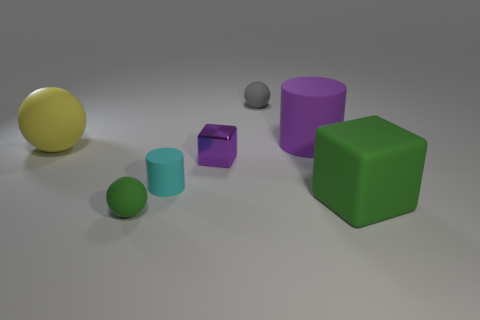}
	\end{minipage}
	 \hspace{1.2cm}
	\begin{minipage}{0.80\linewidth}
		\scriptsize
		\begin{tabular}{ll}
			Human           & what number of other things are there of the same shape as the small purple metallic thing ?  \quad \textbf{A:1} \\
			\midrule
			pretrain        & what number of other objects are the same color as the tiny rubber cylinder ? \\
		    pretrain + RL   & what number of purple objects are either small matte objects or big matte blocks ? \\ \midrule
			scratch         & many gray in big purple purple purple many or many gray matte matte  \\
			scratch+KL-task & what number of other things are the same color as the large rubber cylinder ? \\
			scratch+KL-ext  & how many other things in the are of same color as the large cylinder ?\\ \midrule
			\algo(Task-LM) & how many tiny things have the same color as the large rubber thing ? \\
			\algo(Ext-LM)   & how many other things in the are of the same color as that large thing ? \\
			\bottomrule
		\end{tabular}
	\end{minipage}
	\vspace{1mm}

  \begin{minipage}{0.2\linewidth}
        \includegraphics[width=120px]{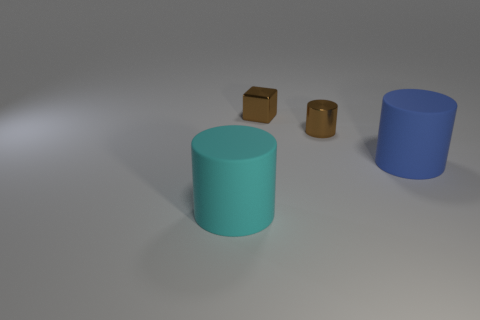}
	\end{minipage}
	\hspace{1.2cm}
	\begin{minipage}{0.80\linewidth}
		\scriptsize
		\begin{tabular}{ll}
			Human           & what shape is the big matte object that is on the right side ofthe big cyan matte object ? \quad \textbf{A:cylinder} \\
			\midrule
			pretrain        & the cyan matte thing that is the same size as the brown object is what shape ? \\
		    pretrain + RL   & what shape is the cyan matte object that is behind the cylinder ? \\ \midrule
			scratch         & many yellow big either either that that that more that metal ?  \\
			scratch+KL-task & what number of other things are the same shape as the small gray thing ? \\
			scratch+KL-ext  & what number of blocks are in the things in the ?\\ \midrule
			\algo(Task-LM) & how many tiny things have the same color as the large rubber thing ? \\
			\algo(Ext-LM)   & what is the shape of that large thing ? \\
			\bottomrule
		\end{tabular}
	\end{minipage}
	\vspace{1mm}
	\captionof{figure}{\small Samples on CLEVR. }
    \label{fig:clevrsamples1}
    \vskip -0.7em
\end{table*}

\begin{table*}[t!]
\label{clevrsamples2}
  \begin{minipage}{0.2\linewidth}
        \includegraphics[width=120px]{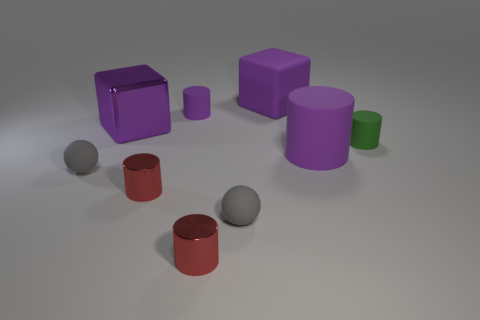}
	\end{minipage}
	\hspace{1.2cm}
	\begin{minipage}{0.80\linewidth}
		\scriptsize
		\begin{tabular}{ll}
			Human           & what is the size of the other rubber cylinder that is the same color as the big cylinder ? \quad \textbf{A:small} \\
			\midrule
			pretrain        & there is a purple object that is the same size as the purple rubber cylinder ; what is its shape? \\
		    pretrain + RL   & what size is the gray ball that is right of the purple sphere ? \\ \midrule
			scratch         & that greater tiny as shiny both are a tiny it either ball right  \\
			scratch+KL-task & there is a big thing that is the same color as the big matte cylinder ; what is its shape? \\
			scratch+KL-ext  & how material is the yellow ?\\ \midrule
			\algo(Task-LM) & how big is the thing that is to the right of the big matte thing ? \\
			\algo(Ext-LM)   & what size is the object that is behind the large red thing ? \\
			\bottomrule
		\end{tabular}
	\end{minipage}
	\vspace{1mm}

  \begin{minipage}{0.2\linewidth}
        \includegraphics[width=120px]{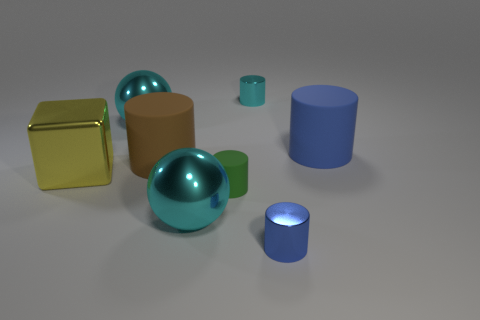}
	\end{minipage}
	\hspace{1.2cm}
	\begin{minipage}{0.80\linewidth}
		\scriptsize
		\begin{tabular}{ll}
			Human           & There is a shiny thing that is both right of the small matte thing and behind the large yellow cube; what size is it ? \quad \textbf{A:small} \\
			\midrule
			pretrain        & there is a big thing that is the same color as the small rubber cylinder ; what is its shape \\
		    pretrain + RL   & there is a brown matte object to the right of the cyan object ; what shape is it ? \\ \midrule
			scratch         & many yellow big either either that that that more that metal ?  \\
			scratch+KL-task & what number of other things are the same shape as the small gray thing ? \\
			scratch+KL-ext  & what is the material of that block ?\\ \midrule
			\algo(Task-LM) & what shape is the big thing that is to the right of the big cyan thing ? \\
			\algo(Ext-LM)   & what is the shape of that large thing ? \\
			\bottomrule
		\end{tabular}
	\end{minipage}
	\vspace{1mm}

  \begin{minipage}{0.2\linewidth}
        \includegraphics[width=120px]{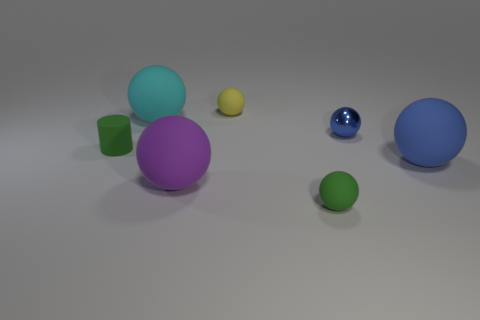}
	\end{minipage}
	\hspace{1.2cm}
	\begin{minipage}{0.80\linewidth}
		\scriptsize
		\begin{tabular}{ll}
			Human           & there is a object that is the same color as the rubber cylinder ;what is its shape ? \quad \textbf{A:sphere} \\
			\midrule
			pretrain        & there is a small thing that is the same color as the small rubber cylinder ; what is its shape? \\
		    pretrain + RL   & what shape is the thing that is the same color as the cylinder ? \\ \midrule
			scratch         & many yellow yellow and cube shape behind cubes shape less small equal shape small equal large large ?  \\
			scratch+KL-task & how many other things in the color are of same material as the green shiny object ? \\
			scratch+KL-ext  & how many spheres anything ?\\ \midrule
			\algo(Task-LM) & what is the shape of the small cyan thing ? \\
			\algo(Ext-LM)   & And shape ? \\
			\bottomrule
		\end{tabular}
	\end{minipage}
	\vspace{1mm}

  \begin{minipage}{0.2\linewidth}
        \includegraphics[width=120px]{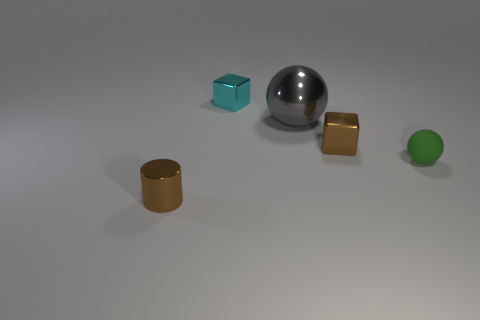}
	\end{minipage}
	\hspace{1.2cm}
	\begin{minipage}{0.80\linewidth}
		\scriptsize
		\begin{tabular}{ll}
			Human           & what is the color of the small thing that is the same shape asthe large gray object ? \quad \textbf{A:green} \\
			\midrule
			pretrain        & there is another rubber object that is the same shape as the small brown object ; what color is it?\\
		    pretrain + RL   & what is the color of the tiny rubber thing ? \\ \midrule
			scratch         & many sphere less how an  \\
			scratch+KL-task & what number of other objects are the same shape as the tiny blue object ? \\
			scratch+KL-ext  & what size is that cylinder ?\\ \midrule
			\algo(Task-LM) & what is the color of the tiny matte thing ? \\
			\algo(Ext-LM)   & what color is the small thing ? \\
			\bottomrule
		\end{tabular}
	\end{minipage}
	\vspace{1mm}
	
  \begin{minipage}{0.2\linewidth}
        \includegraphics[width=120px]{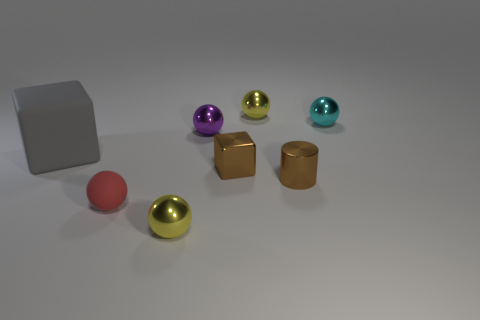}
	\end{minipage}
	\hspace{1.2cm}
	\begin{minipage}{0.80\linewidth}
		\scriptsize
		\begin{tabular}{ll}
			Human           & what number of shiny objects are cyan spheres or tiny balls ? \quad \textbf{A:4} \\
			\midrule
			pretrain        & how many other things are there of the same size as the brown rubber thing ?\\
		    pretrain + RL   & how many other things are there of the same size as the cyan rubber thing ? \\ \midrule
			scratch         & many yellow do do do either do either do balls  \\
			scratch+KL-task & what number of other things are there of the same shape as the small gray thing ? \\
			scratch+KL-ext  & how many other things are of same color as ball ?\\ \midrule
			\algo(Task-LM) & how many other things are the same material as the small cyan cylinder ? \\
			\algo(Ext-LM)   & how many other things in the material of the small thing that is the same material as green thing ? \\
			\bottomrule
		\end{tabular}
	\end{minipage}
	\vspace{1mm}

	\captionof{figure}{\small Samples on Clevr. }
    \label{fig:clevrsamples2}
    \vskip -0.7em
\end{table*}

\begin{table*}[t!]
    \begin{minipage}{0.2\linewidth}
        \includegraphics[width=120px]{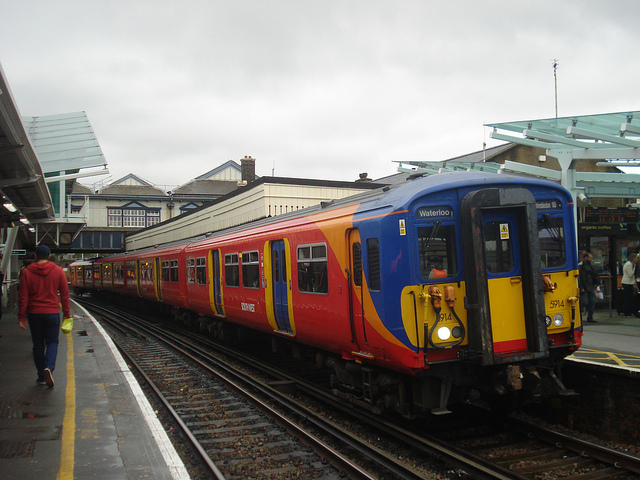}
	\end{minipage}
	\hspace{1.2cm}
	\begin{minipage}{0.80\linewidth}
		\scriptsize
		\begin{tabular}{ll}
			Human           & How many trains?  \quad \textbf{A:1} \\
			\midrule
			pretrain        & How many trains are in the picture? \\
		    pretrain + RL   & How many trains are shown? \\ \midrule
			scratch         & AmazingAmazingAmazingAmazingAmazingAmazing  \\
			scratch+KL-task & How many people are in the picture? \\
			scratch+KL-ext  & The the same way of the same way of the\\ \midrule
			\algo(Task-LM) & How many windows are here? \\
			\algo(Ext-LM)   & How many different ways would we take them to reach \\
			\bottomrule
		\end{tabular}
	\end{minipage}
	\vspace{1mm}
\begin{minipage}{0.2\linewidth}
		\centering
        \includegraphics[width=120px]{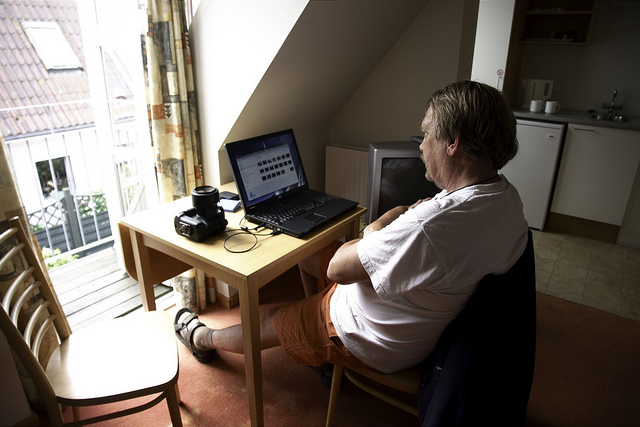}
	\end{minipage}
	\hspace{1.2cm}
	\begin{minipage}{0.85\linewidth}
	\small
		\scriptsize
		\begin{tabular}{ll}
			\toprule
			Human            & What is the man wearing over his shirt?  \quad \textbf{A:nothing} \\
			\midrule
			pretrain        &  What is in front of the person? \\
		    pretrain + RL   & What is written on the person's right?  \\ \midrule
			scratch         & AmazingAmazingAmazingAmazingAmazing \\
			scratch+KL-task & Is this child old or unloaded? \\
			scratch+KL-ext  & The the first thing of the world’s a few \\ \midrule
			\algo(Task-LM)  & How many different shapes have been used and used by \\
			\algo(Ext-LM)   & How many planes can be seen on top? \\
			\bottomrule
		\end{tabular}
	\end{minipage}
	\vspace{1mm}
\begin{minipage}{0.2\linewidth}
		\centering
        \includegraphics[width=120px]{./COCO_val2014_000000090003.jpg}
	\end{minipage}
	\hspace{1.2cm}
	\begin{minipage}{0.85\linewidth}
	\small
		\scriptsize
		\begin{tabular}{ll}
			\toprule
			Human            & Is the fence in the background metal or wood?  \quad \textbf{A:metal} \\
			\midrule
			pretrain        &  What kind of dog is on the dog? \\
		    pretrain + RL   & What kind of dog is this?  \\ \midrule
			scratch         & AmazingAmazingAmazingAmazingAmazing \\
			scratch+KL-task & What is 150 seated over theware? \\
			scratch+KL-ext  & The the same way of the same way to be \\ \midrule
			\algo(Task-LM)  & Which item has more aggressive up their legs? \\
			\algo(Ext-LM)   & How many different species have been captured and tagged here \\
			\bottomrule
		\end{tabular}
	\end{minipage}
	
\begin{minipage}{0.2\linewidth}
		\centering
        \includegraphics[width=120px]{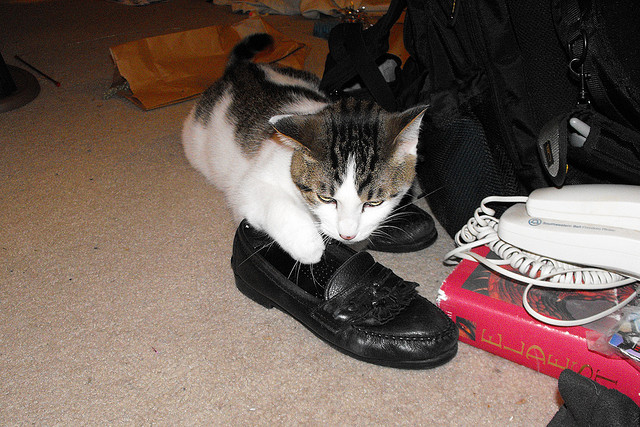}
	\end{minipage}
	\hspace{1.2cm}
	\begin{minipage}{0.85\linewidth}
	\small
		\scriptsize
		\begin{tabular}{ll}
			\toprule
			Human            & What is the title of the red book?  \quad \textbf{A:harry potter} \\
			\midrule
			pretrain        &  What is the cat inside of? \\
		    pretrain + RL   & Is the cat inside or outside?  \\ \midrule
			scratch         & AmazingAmazingAmazingAmazingAmazing \\
			scratch+KL-task & What color is the man's shirt? \\
			scratch+KL-ext  & The way of the world’s a lot of the \\ \midrule
			\algo(Task-LM)  & Which item appears higher into one and lower? \\
			\algo(Ext-LM)   & What was that? \\
			\bottomrule
		\end{tabular}
	\end{minipage}
	
\begin{minipage}{0.2\linewidth}
		\centering
        \includegraphics[width=120px]{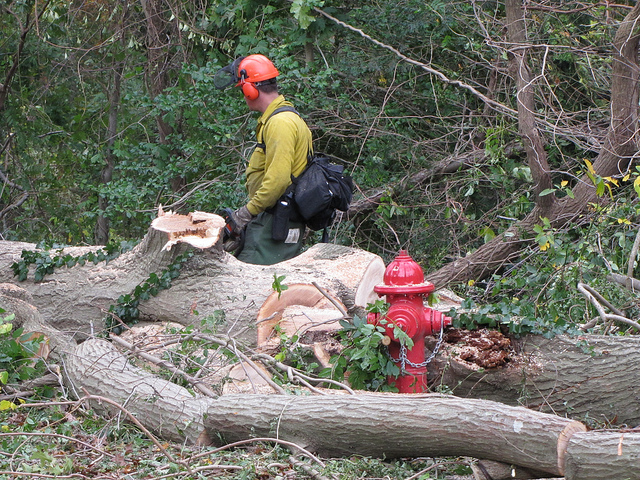}
	\end{minipage}
	\hspace{1.2cm}
	\begin{minipage}{0.85\linewidth}
	\small
		\scriptsize
		\begin{tabular}{ll}
			\toprule
			Human            & What color is the fire hydrant?  \quad \textbf{A:red} \\
			\midrule
			pretrain        &  What color is the hydrant? \\
		    pretrain + RL   & What color is the hydrant?  \\ \midrule
			scratch         & AmazingAmazingAmazingAmazingAmazing \\
			scratch+KL-task & What color is the man's bat? \\
			scratch+KL-ext  & The the first thing is a good thing that the \\ \midrule
			\algo(Task-LM)  & Which color is this fire? \\
			\algo(Ext-LM)   & What color will your feet color look? \\
			\bottomrule
		\end{tabular}
	\end{minipage}
	
\begin{minipage}{0.2\linewidth}
		\centering
        \includegraphics[width=120px]{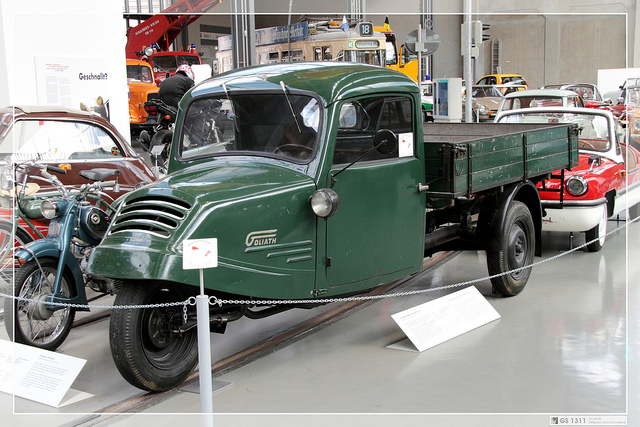}
	\end{minipage}
	\hspace{1.2cm}
	\begin{minipage}{0.85\linewidth}
	\small
		\scriptsize
		\begin{tabular}{ll}
			\toprule
			Human            & How many wheels does the truck have?  \quad \textbf{A:3} \\
			\midrule
			pretrain        &  How many people are in front of the bus? \\
		    pretrain + RL   & How many slices ofists are on the plate?  \\ \midrule
			scratch         & AmazingAmazingAmazingAmazingAmazing \\
			scratch+KL-task & Is summer out or cloudy next to Winchester? \\
			scratch+KL-ext  & The the most recent of the most recent years of \\ \midrule
			\algo(Task-LM)  & How many pieces are here? \\
			\algo(Ext-LM)   & How many different objects have been used? \\
			\bottomrule
		\end{tabular}
	\end{minipage}	
	\captionof{figure}{\small Samples on VQA. }
    \label{fig:vqasamples1}
    \vskip -0.7em
\end{table*}

\begin{table*}[t!]
    \begin{minipage}{0.2\linewidth}
        \includegraphics[width=120px]{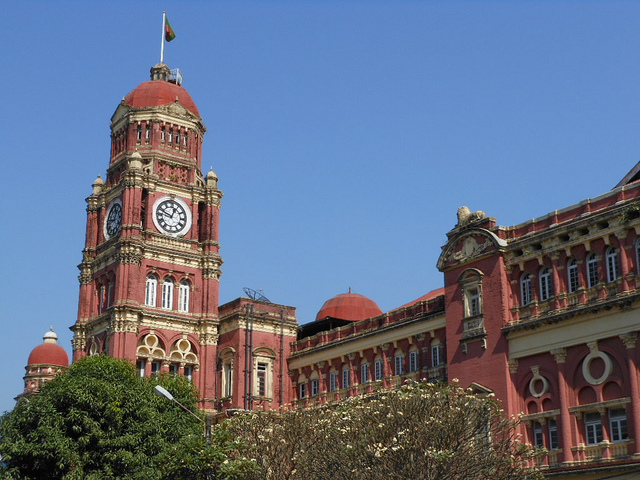}
	\end{minipage}
	\hspace{1.2cm}
	\begin{minipage}{0.80\linewidth}
		\scriptsize
		\begin{tabular}{ll}
			Human           & What is on top of the round dome?  \quad \textbf{A:flag} \\
			\midrule
			pretrain        & What is on the top right mean? \\
		    pretrain + RL   & What is on the front of this event? \\ \midrule
			scratch         & AmazingAmazingAmazingAmazingAmazing  \\
			scratch+KL-task & What zombie is on the mouse? \\
			scratch+KL-ext  & The the first thing is a bit of the first\\ \midrule
			\algo(Task-LM) & Where could one travel park located? \\
			\algo(Ext-LM)   & What color will your shoes look? \\
			\bottomrule
		\end{tabular}
	\end{minipage}
	\vspace{1mm}

    \begin{minipage}{0.2\linewidth}
        \includegraphics[height= 110px]{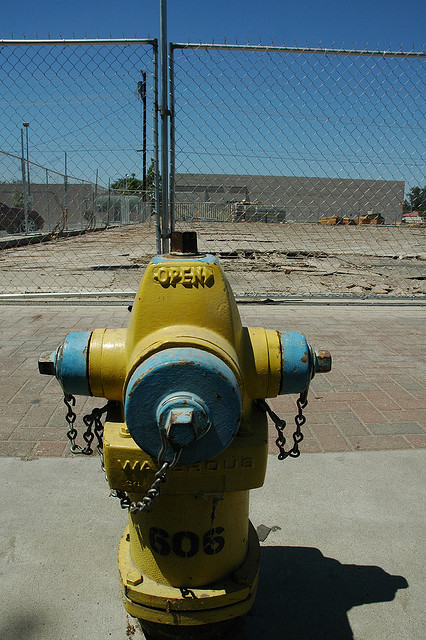}
	\end{minipage}
	\hfill
	\begin{minipage}{0.80\linewidth}
		\scriptsize
		\begin{tabular}{ll}
			Human           & Where is the chain link fence?  \quad \textbf{A:background} \\
			\midrule
			pretrain        & What is he holding? \\
		    pretrain + RL   & What is he fire hydrant? \\ \midrule
			scratch         & mazingAmazingAmazingAmazingAmazing  \\
			scratch+KL-task & Who is closest to the paint? \\
			scratch+KL-ext  & The the first thing is a great deal with the\\ \midrule
			\algo(Task-LM) & Which item represents than both ends and lower? \\
			\algo(Ext-LM)   & How much food has it given him? \\
			\bottomrule
		\end{tabular}
	\end{minipage}
	\vspace{1mm}
	
   \begin{minipage}{0.2\linewidth}
        \includegraphics[width=120px]{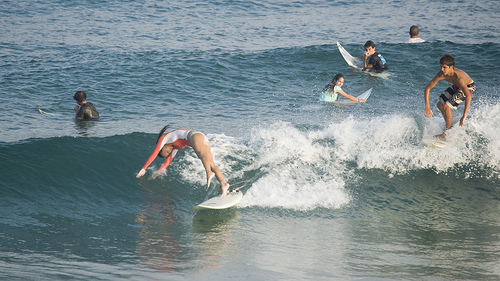}
	\end{minipage}
	\hspace{1.2cm}
	\begin{minipage}{0.80\linewidth}
		\scriptsize
		\begin{tabular}{ll}
			Human           & What activity are these people doing?  \quad \textbf{A:surfing} \\
			\midrule
			pretrain        & What is the person doing? \\
		    pretrain + RL   & What is the person doing? \\ \midrule
			scratch         & noodles noodles noodles noodles noodles noodles  \\
			scratch+KL-task & How many umbrellas are visible? \\
			scratch+KL-ext  & The the first thing is the same way of the\\ \midrule
			\algo(Task-LM) & Which game does he play? \\
			\algo(Ext-LM)   & What was that for? \\
			\bottomrule
		\end{tabular}
	\end{minipage}
	\vspace{1mm}

   \begin{minipage}{0.2\linewidth}
        \includegraphics[width=120px]{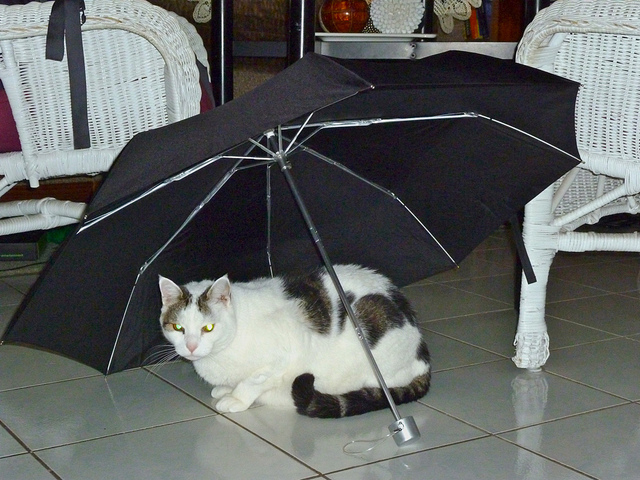}
	\end{minipage}
	\hspace{1.2cm}
	\begin{minipage}{0.80\linewidth}
		\scriptsize
		\begin{tabular}{ll}
			Human           & What color is the umbrella?  \quad \textbf{A:black} \\
			\midrule
			pretrain        & What color is the cat? \\
		    pretrain + RL   & What color is the cat? \\ \midrule
			scratch         & AmazingAmazingAmazingAmazingAmazing  \\
			scratch+KL-task & What color is the man's shirt? \\
			scratch+KL-ext  & The the other way of the past time, and\\ \midrule
			\algo(Task-LM) & Which item doesn't both turn? \\
			\algo(Ext-LM)   & What color of clothing did he get? \\
			\bottomrule
		\end{tabular}
	\end{minipage}
	\vspace{1mm}

   \begin{minipage}{0.2\linewidth}
        \includegraphics[width=120px]{./COCO_val2014_000000216115.jpeg}
	\end{minipage}
	\hspace{1.2cm}
	\begin{minipage}{0.80\linewidth}
		\scriptsize
		\begin{tabular}{ll}
			Human           & How many planes are shown?  \quad \textbf{A:1} \\
			\midrule
			pretrain        & How many jets are there? \\
		    pretrain + RL   & How many jets are there? \\ \midrule
			scratch         & AmazingAmazingAmazingAmazingAmazingAmazing  \\
			scratch+KL-task & How many skater does Green cents have? \\
			scratch+KL-ext  & The the first thing is the first time, and\\ \midrule
			\algo(Task-LM) & How many surf worthy are here? \\
			\algo(Ext-LM)   & How many different ways should one ask if she wants \\
			\bottomrule
		\end{tabular}
	\end{minipage}
	\vspace{1mm}

   \begin{minipage}{0.2\linewidth}
        \includegraphics[height=110px]{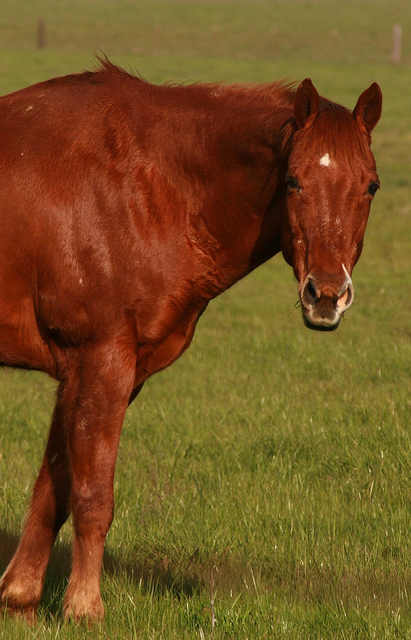}
	\end{minipage}
	\hfill
	\begin{minipage}{0.80\linewidth}
		\scriptsize
		\begin{tabular}{ll}
			Human           & What is this animal called?  \quad \textbf{A:horse} \\
			\midrule
			pretrain        & What is the animal on? \\
		    pretrain + RL   & What animal is shown on the ground? \\ \midrule
			scratch         & AmazingAmazingAmazingAmazingAmazingAmazingAmazing  \\
			scratch+KL-task & What has to make of the pies that, should \\
			scratch+KL-ext  & The the next week of the next week, the\\ \midrule
			\algo(Task-LM) & Which item doesn't turn? \\
			\algo(Ext-LM)   & What was that? \\
			\bottomrule
		\end{tabular}
	\end{minipage}
	\vspace{1mm}
	\captionof{figure}{\small Samples on VQA. }
    \label{fig:vqasamples2}
    \vskip -0.7em
\end{table*}

\begin{table*}[t!]
   \begin{minipage}{0.2\linewidth}
        \includegraphics[height=110px]{./COCO_val2014_000000275025.jpeg}
	\end{minipage}
	\hfill
	\begin{minipage}{0.80\linewidth}
		\scriptsize
		\begin{tabular}{ll}
			Human           & What color spot does the horse have?  \quad \textbf{A:white} \\
			\midrule
			pretrain        & What color is the animal? \\
		    pretrain + RL   & What color is the door? \\ \midrule
			scratch         & AmazingAmazingAmazingAmazingAmazingAmazing  \\
			scratch+KL-task & What color is the ATM basketball? \\
			scratch+KL-ext  & The the same thing that the same way of the\\ \midrule
			\algo(Task-LM) & Which color is his socks? \\
			\algo(Ext-LM)   & What color will your shoes look? \\
			\bottomrule
		\end{tabular}
	\end{minipage}
	\vspace{1mm}

   \begin{minipage}{0.2\linewidth}
        \includegraphics[width=120px]{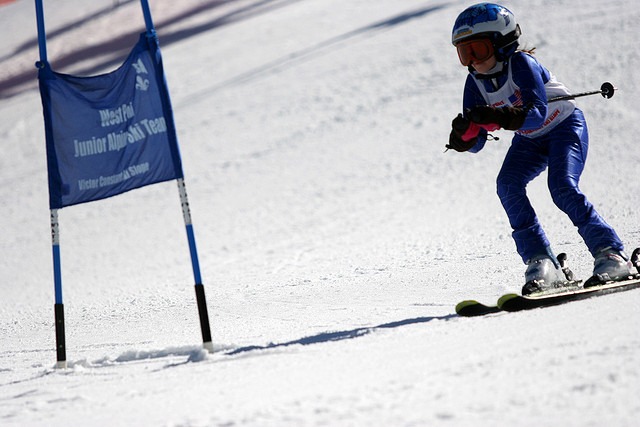}
	\end{minipage}
	\hspace{1.2cm}
	\begin{minipage}{0.80\linewidth}
		\scriptsize
		\begin{tabular}{ll}
			Human           & What color is the girls pants?  \quad \textbf{A:blue} \\
			\midrule
			pretrain        & What color is the man's blue? \\
		    pretrain + RL   & What color are the bird's pants? \\ \midrule
			scratch         & AmazingAmazingAmazingAmazingAmazingAmazing \\
			scratch+KL-ext  & The the first thing is a lot of the same\\ \midrule
			\algo(Task-LM) & Which color is this fire? \\
			\algo(Ext-LM)   & What color of clothing did he get? \\
			\bottomrule
		\end{tabular}
	\end{minipage}
	\vspace{1mm}

   \begin{minipage}{0.2\linewidth}
        \includegraphics[width=120px]{./COCO_val2014_000000275034.jpeg}
	\end{minipage}
	\hspace{1.2cm}
	\begin{minipage}{0.80\linewidth}
		\scriptsize
		\begin{tabular}{ll}
			Human           & What is on the woman's head?  \quad \textbf{A:helmet} \\
			\midrule
			pretrain        & What is on the girl's head? \\
		    pretrain + RL   & What is on the person's head? \\ \midrule
			scratch         & AmazingAmazingAmazingAmazingAmazingAmazing  \\
			scratch+KL-task & Who is behind the horse? \\
			scratch+KL-ext  & The the same thing that the most important to the\\ \midrule
			\algo(Task-LM) & Which item doesn't turn? \\
			\algo(Ext-LM)   & What was that? \\
			\bottomrule
		\end{tabular}
	\end{minipage}
	\vspace{1mm}

	\captionof{figure}{\small Samples on VQA. }
    \label{fig:vqasamples3}
    \vskip -0.7em
\end{table*}

\end{document}